\documentclass{article}

\usepackage{microtype}
\usepackage{graphicx}
\usepackage{subfigure}
\usepackage{booktabs} 

\usepackage{hyperref}

\usepackage[accepted]{icml2026}

\usepackage{amsmath}
\usepackage{amssymb}
\usepackage{mathtools}
\usepackage{amsthm}

\usepackage{booktabs}
\usepackage{multirow}
\usepackage{multicol}
\usepackage{arydshln}
\usepackage{colortbl}
\usepackage{xcolor}

\usepackage[capitalize,noabbrev]{cleveref}

\theoremstyle{plain}

\theoremstyle{definition}

\theoremstyle{remark}

\usepackage[textsize=tiny]{todonotes}

\icmltitlerunning{Unsupervised Synthetic Image Attribution: Alignment and Disentanglement}

\begin{document}

\twocolumn[
\icmltitle{Unsupervised Synthetic Image Attribution: Alignment and Disentanglement}



\icmlsetsymbol{equal}{*}

\begin{icmlauthorlist}
\icmlauthor{Zongfang Liu}{equal,mbz}
\icmlauthor{Guangyi Chen}{equal,cmu,mbz}
\icmlauthor{Boyang Sun}{mbz}
\icmlauthor{Tongliang Liu}{usyd,mbz}
\icmlauthor{Kun Zhang}{cmu,mbz}
\end{icmlauthorlist}

\icmlaffiliation{mbz}{Mohamed bin Zayed University of Artificial Intelligence}
\icmlaffiliation{cmu}{Carnegie Mellon University}
\icmlaffiliation{usyd}{The University of Sydney}

\icmlcorrespondingauthor{Kun Zhang}{kunz1@cmu.edu}

\icmlkeywords{Machine Learning, ICML}

\vskip 0.3in
]



\printAffiliationsAndNotice{\icmlEqualContribution} 

\begin{abstract}
As the quality of synthetic images improves, identifying the underlying concepts of model-generated images is becoming increasingly crucial for copyright protection and ensuring model transparency. Existing methods achieve this attribution goal by training models using annotated pairs of synthetic images and their original training sources. However, obtaining such paired supervision is challenging, as it requires either well-designed synthetic concepts or precise annotations from millions of training sources.
To eliminate the need for costly paired annotations, in this paper, we explore the possibility of unsupervised synthetic image attribution. We propose a simple yet effective unsupervised method called Alignment and Disentanglement. Specifically, we begin by performing basic concept alignment using contrastive self-supervised learning. Next, we enhance the model's attribution ability by promoting representation disentanglement with the Infomax loss. 
This approach is motivated by an interesting observation: contrastive self-supervised models, such as MoCo and DINO, inherently exhibit the ability to perform simple cross-domain alignment. By formulating this observation as a theoretical assumption on cross-covariance, we provide a theoretical explanation of how alignment and disentanglement can approximate the concept-matching process through a decomposition of the canonical correlation analysis objective.
On the real-world benchmarks, AbC, we show that our unsupervised method surprisingly outperforms the supervised methods. As a starting point, we expect our intuitive insights and experimental findings to provide a fresh perspective on this challenging task.
\end{abstract}

\section{Introduction}

Generative models are capable of producing high-quality synthetic images distinct from those in their training sets by merging elements from various images to create new scenes. Given the complexities and necessities of copyright and ownership, the attribution of synthetic images is critical. This process requires a deep understanding of the connections between the training data and the outputs generated by the model. It is essential to ensure that original content creators are properly recognized and compensated, address legal and ethical issues, aid in the detection of data leaks, and trace the origins of content.
\begin{figure}
    \centering
    \includegraphics[width=0.45\textwidth]{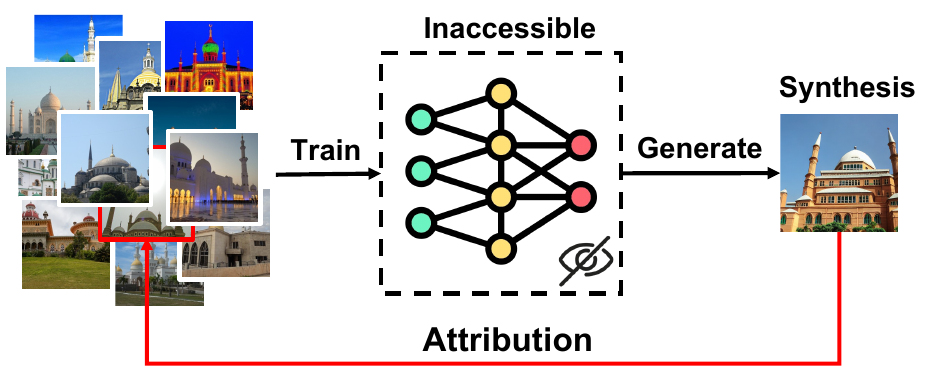}
    \caption{\textbf{Synthetic image attribution without model access.} Modern generative models are often closed-source and trained on vast datasets, making it challenging to attribute a generated image to its true training exemplar(s) when the model is inaccessible.}
    \label{fig:attribution_illustration}
    \vspace{-0.6cm}
\end{figure}

Despite its importance, the task of data attribution remains a significant challenge and an open issue within the community, particularly for large-scale generative models. One promising method is to analyze the individual effects of each training sample. For instance, influence functions \cite{law1986robust, huber2011robust, koh2017understanding} are employed to assess the impact of specific training examples on a model’s predictions, identifying the most influential data points. Similarly, Shapley value estimators \cite{feldman2020neural} distribute credit to each data point based on its contribution to the model’s overall performance. However, these methods struggle to scale up to the demands of current generative models \cite{ramesh2022hierarchical, saharia2022photorealistic, rombach2022high}, which involve training with millions or even billions of images. 
Recently, TRAK~\cite{park2023trak} introduced a kernel machine-based approximation that enables faster attribution scoring in a surrogate space.
Subsequent work adapted these ideas to diffusion models, including D-TRAK~\cite{zheng2023intriguing} and Journey-TRAK~\cite{georgiev2023journey}, as well as more direct diffusion-specific formulations~\cite{mlodozeniec2024influence,lin2024diffusion}.
In parallel, recent efforts have also started to target text-to-image settings and improve practical efficiency~\cite{wang2025fast}. Despite this progress, scaling attribution to modern generative models remains difficult. Beyond sheer dataset and model size, many existing methods still presuppose access to model internals or training-time artifacts (e.g., parameters, gradients, checkpoints, or logs), which is often unrealistic for proprietary or closed-source deployments. This has motivated emerging directions that reduce reliance on such access, including model-agnostic or data-only formulations tailored to diffusion models~\cite{zhao2025nonparametric}.

Another potential approach involves learning a latent representation space to align the concepts of synthetic images with their original training images. However, existing methods \cite{wang2023evaluating} for learning such a latent space typically require paired data for supervision. Obtaining this paired supervision is non-trivial; it requires either the creation of well-designed, customized synthetic concepts \cite{wang2023evaluating} or the extraction of ground-truth training samples from millions of potential sources \cite{feldman2020neural}. This computationally intensive annotation process exhibits limited generalization capabilities, and motivates us to think about the following important yet unresolved question:

\begin{center}
	{ \textit{Is it possible to learn the concept matching space without } } \\
	{ \textit{paired annotation for synthetic image attribution?} }
\end{center}

In this paper, we explore this question and propose a method, called Alignment and Disentanglement. During the Alignment phase, we apply a contrastive self-supervised learning method to achieve basic concept alignment by a shared mapping, targeting a latent space that allows comparison between synthetic and original training images.
This method is inspired by experimental findings that demonstrate the inherent capability of models like MOCO \cite{he2020momentum} and DINO \cite{caron2021emerging} to achieve cross-domain alignment.
In the Disentanglement phase, we aim to further bridge the domain gap using Independent Component Analysis (ICA). Specifically, we learn two domain-specific models utilizing the Infomax loss~\cite{bell1995information} to ensure that the latent variables are distributed within the same component-independent multivariate distribution. However, the independent ICA optimizations on two domains may cause permutations of the latent variables. To address these issues, we introduce a regularization term along with identity initialization for the mapping matrix to penalize potential permutations.

Given that the task of matching concepts across synthetic and original domains aligns with a canonical correlation analysis (CCA) framework \cite{hotelling1992relations}, we provide a theoretical explanation of how our method approximates this concept-matching process by decomposing the CCA targets. We discovered that the processes of alignment and disentanglement effectively serve as trivial solutions for CCA, optimizing across and within domains, respectively.
In our experiments, we evaluate our methods using AbC \cite{wang2023evaluating}, a large-scale benchmark renowned for its customized concept generation for synthetic image attribution. Our Alignment and Disentanglement method effectively eliminates the need for paired data and achieves performance comparable to, and at times even surpassing, that of supervised methods.

Our contributions are summarized as follows: 
\begin{itemize}
\vspace{-0.23cm}
\item {We introduce a new unsupervised framework for synthetic image attribution that incorporates alignment and disentanglement.}
\vspace{-0.23cm}
\item We provide a theoretical foundation for our unsupervised framework through the lens of CCA.
\vspace{-0.23cm}
\item We evaluate our framework on AbC, a large-scale benchmark with $>$4M images spanning object and style concepts, and report results on four test datasets covering both in-distribution and out-of-distribution settings; our method achieves performance comparable to supervised approaches even without paired data.
\end{itemize}

\section{Related Work}

\subsection{Data Attribution}
Currently, the majority of research on data attribution problems is based on influence analysis methods, which involve quantifying the impact of individual data points on the predictions of a machine learning model, helping to understand how specific examples affect model behavior and performance. 
Retraining-based influence analysis methods~\cite{ghorbani2019data, jia2019efficient, jia2021scalability, kwon2021beta, kandpal2022deduplicating} directly measure how the inclusion or exclusion of specific data points affects the model’s performance. Gradient-based influence analysis methods~\cite{sharchilev2018finding, koh2019accuracy, schioppa2022scaling, park2023trak} estimate influence by analyzing the gradients of the loss function with respect to the model parameters. 
Based on similar paradigms, recent works~\cite{dai2023training, zheng2023intriguing, georgiev2023journey, wang2024data, mlodozeniec2024influence, lin2024diffusion} provide solutions for data attribution in diffusion models.  
While effective, many influence-based methods require training-time access or model internals, which is often unavailable in practice; recent work improves efficiency for text-to-image attribution~\cite{wang2025fast} and explores more model-agnostic/data-only formulations for diffusion models~\cite{zhao2025nonparametric}. \cite{wang2023evaluating} provides a supervised method that considers the data attribution problem for generative models as an image retrieval task. Though their method is practical because it does not require access to the generative model, it relies on paired synthetic--exemplar data, which is resource-intensive to acquire. To mitigate this issue, we aim to learn an attribution latent space without relying on paired annotations.

\subsection{Canonical Correlation Analysis}
In the supervised learning paradigm, learning the latent space to match the data from two different views can be formulated as Canonical Correlation Analysis (CCA) \cite{hotelling1992relations,bach2005probabilistic}, which finds the maximum projection for two sets of random vectors. It can be viewed as finding the shared latent variables from the observation in different generation function, as shown in the probabilistic interpretation \cite{bach2005probabilistic}. 
Methods like \cite{bach2002kernel, michaeli2016nonparametric, lindenbaum2020multi} extend CCA to the non-linear case by the kernel trick. Deep CCA \cite{andrew2013deep} and some following work ~\cite{lu2015deep,wang2015unsupervised}  use neural networks to learn the non-parametric nonlinear mapping. While CCA is inherently well-suited for cross-domain alignment, traditional CCA methods rely on paired data, limiting their applicability in scenarios where paired data is infeasible. Building on \cite{bach2005probabilistic} our work offers a theoretical analysis and provides a potential solution to enable CCA without paired data.

\subsection{Disentanglement Representation Learning}
Unsupervised (self-supervised) representation learning \cite{radford2021learning, caron2021emerging, chen2020simple, he2020momentum, tian2020contrastive, misra2020self} has shown considerable success in developing powerful and generalizable representations, but often lacks explainability. To enhance explainability and model controllability, methods have been developed to disentangle these representations~\cite{chen2018isolating, zheng2019disentangling, burgess2018understanding,denton2017unsupervised,higgins2018towards,lee2018diverse}.
Within these methods, those based on Independent Component Analysis (ICA) demonstrate strong identifiability properties for latent representations with better theoretical disentanglement guarantees.  
In the early stage, linear ICA~\cite{comon1994independent, bell1995information, hyvarinen2001independent} requires assuming the non-Gaussian latent distribution. One of the widely used methods to achieve linear ICA is InfoMax loss~\cite{bell1995information}, which is applied in our implementation due to its flexibility. Recently, nonlinear ICA \cite{hyvarinen2019nonlinear, sorrenson2020disentanglement, halva2020hidden, lachapelle2022disentanglement} have obtained robust theoretical results supporting the identifiability of latent variables, and enabled the use of deep neural networks such as VAE~\cite{kingma2013auto} to address complex scenarios.

\section{Method}
\begin{figure}
    \centering
    \includegraphics[width=0.5\textwidth]{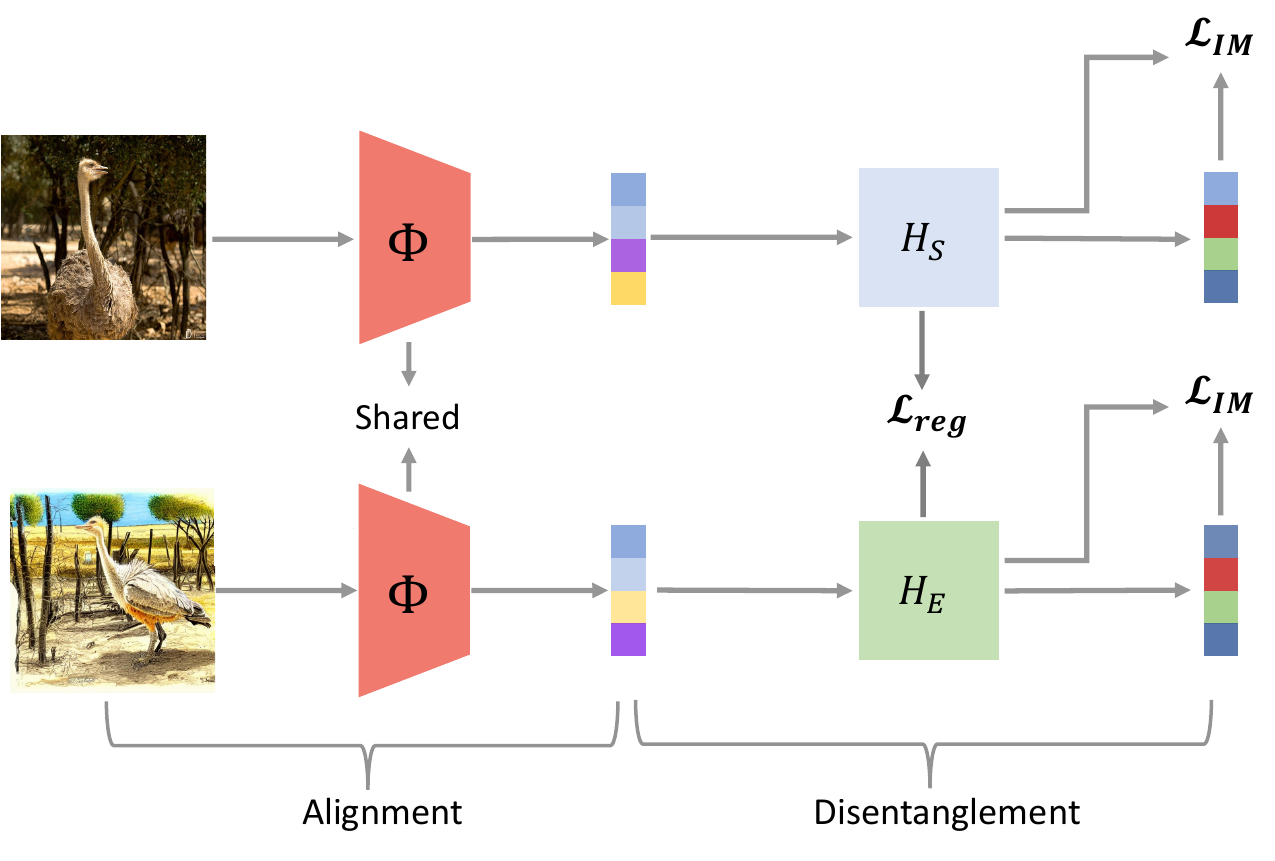}
    \caption{\textbf{Framework of Alignment and Disentanglement.} We begin by extracting roughly aligned features using a shared pretrained backbone ($\mathbf{\Phi}$) through self-supervised contrastive learning. These features are then refined to be more independent and identifiable using two distinct linear mappings, $H_S$ and $H_E$, optimized with Infomax loss (ICA). To support cross-domain alignment and prevent harmful permutations introduced by ICA, we apply identity initialization and orthogonal constraints. This framework effectively functions as a CCA method without paired annotations.}
    \label{fig:model-pipeline}
    \vspace{-0.6cm}
\end{figure}
In this section, we define the problem of synthetic image attribution and present our unsupervised framework, Alignment and Disentanglement (A\&D), which identifies the source images behind synthetic outputs without the need for paired annotations. Our approach proceeds in two main stages: alignment, where contrastive self-supervised learning establishes an initial cross-domain similarity, and disentanglement, using Independent Component Analysis (ICA) to make learned features more independent and identifiable, thereby enhancing alignment. Furthermore, we provide a theoretical analysis of our A\&D approach from a Canonical Correlation Analysis (CCA) perspective, specifying conditions that enable CCA to operate without paired data.
\subsection{Problem: Synthetic Image Attribution}
Synthetic image attribution focuses on identifying the source training image behind model-generated synthetics. In this paper, we focus on the challenging scenario described in \cite{wang2023evaluating}, where only the training and synthetic images are accessible, without touching the model parameters. This setup is more realistic, as more generative models are becoming closed-source. Formally, consider two sets of images: $X_{S}=\{\mathbf{x}_{S}^{(i)}\}_{i=1}^N \subset \mathcal{X}_{S}$, which represents the synthetic images, and $X_{E}=\{\mathbf{x}^{(j)}_{E}\}_{j=1}^M \subset \mathcal{X}_{E}$, which represents the original training (exemplar) data. For convenience and aligning with \cite{wang2023evaluating}, we use exemplar data/domain to represent the original training data in the following sections. 
In the supervised setting, the annotated dataset with synthesis-exemplar pairs $\mathcal{D}_{paired}= \{(\mathbf{x}^{(k)}_{S}, \mathbf{x}^{(k)}_{E}\}_{k=1}^K$ are given. In our unsupervised setting, there is no paired data and only $X_{S}$ and $X_{E}$ are given.

The goal of synthetic image attribution is to learn two mapping functions, $F_S: \mathbf{x}_{S} \rightarrow \mathbf{z}$ for the synthetic domain and $F_E: \mathbf{x}_{E} \rightarrow \mathbf{z}$ for the exemplar domain, respectively. This allows us to match concepts and perform retrieval to identify the corresponding exemplar for a given synthetic image.

\subsection{Alignment and Disentanglement}

As the name suggests, our approach consists of two key stages: alignment through contrastive self-supervised learning and disentanglement using ICA.  
The overall framework is illustrated in Figure~\ref{fig:model-pipeline}.

\textbf{Alignment.} The alignment is achieved by contrastive self-supervised learning, whose basic idea is optimizing embeddings so that positive pairs are closer and negative pairs are farther apart in a feature space. Formally, the objective function can be formulated as 
\begin{equation}
\label{eq:generation}
\begin{aligned}
\mathcal{L}_{\text{C}}^{i,j} = - \log \frac{\exp(\text{sim}(\mathbf{z}^{(i)},\mathbf{z}^{(j)}) / \tau)}{\sum_k \mathbb{I}_{[k\neq i]} \exp(\text{sim}(\mathbf{z}^{(i)},\mathbf{z}^{(k)}) / \tau)}, 
\end{aligned}
\end{equation}
where, $\text{sim}(\cdot,\cdot)$ denotes the similarity metric, such as cosine similarity. Here, $\mathbf{z}^{(i)}$ and $\mathbf{z}^{(j)}$ denote the representations of the query and positive samples, respectively, which are typically generated by different augmentations. As the specifics of conducting contrastive learning are not the core contribution of this framework, we do not elaborate extensively here. We refer the reader to the references~\cite{he2020momentum,chen2020simple,caron2021emerging} for more details.

Instead, we focus on explaining why self-supervised learning methods can achieve alignment. Leveraging the inherent data structure information, the alignment process highlights features that distinguish a sample from other ``negative'' samples. Despite the persistence of shifted domain information, this process establishes a basic feature space that facilitates rough matching between synthetics and exemplars. In our experiments, we observed that: 1) it is feasible to use pre-trained models without fine-tuning on the exemplars, and 2) with conservative training, the alignment progressively improves (shown in \cref{tab:DINO-tuned}).

\textbf{Disentanglement.}
After alignment, we apply ICA to further refine the representations of two domains, ensuring they conform to the same component-independent multivariate distribution. In particular, we introduce two distinct linear mappings, $\mathbf{H}_{S}: \mathbf{z}_{S} \in \mathbb{R}^{D} \rightarrow \tilde{\mathbf{z}}_{S} \in \mathbb{R}^{D}$ and $\mathbf{H}_{E}: \mathbf{z}_{E} \in \mathbb{R}^{D} \rightarrow \tilde{\mathbf{z}}_{E} \in \mathbb{R}^{D}$, for the synthetic and exemplar domains, respectively. 
 $D$ is the dimension of the feature before and after ICA mapping. Each domain is fine-tuned using the objective function derived from the Infomax ICA principle \cite{bell1995information}. For instance, in the synthetic domain, the Infomax loss function is expressed as:
\begin{equation}
\mathcal{L}_{\text{IM}} (X_S) = \ln |\mathbf{H}_{S}| + \mu \mathcal{L}_{\text{Ent}},
\end{equation}
where $\mathbf{H}_{S} \in \mathbb{R}^{D \times D} $ denotes the weight matrix of the linear mapping. 
$\mathcal{L}_{\text{Ent}}$ calculates the entropy of the latent variables: 
\begin{equation}
\mathcal{L}_{\text{Ent}} = \frac{1}{ND} \sum_{i=1}^{N} \sum_{d=1}^{D} \ln\left(1 - \tanh\left(\mathbf{h}_{S,d}^\top \mathbf{z}_{S}^{(i)}\right)^2\right) ,
\end{equation}
 where $\mathbf{h}_{S,d}$ is the $d$th column of the mapping matrix $\mathbf{H}_{S}$ and $\mathbf{z}_{S}^{(i)}$ denotes the feature of $i$th sample obtained by the alignment process. Then, $\tilde{\mathbf{z}}_{S}^{d} = \mathbf{h}_{S,d}^\top \mathbf{z}_{S}^{(i)}$ produces the $l$th variable. $\mu$ denotes the ratio of the entropy term.

\begin{figure}
    \centering
    \includegraphics[width=0.4\linewidth]{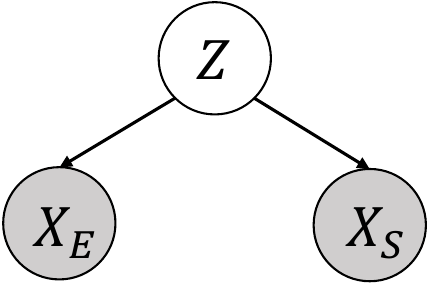}
    \caption{\textbf{Data generation process for CCA.} A shared latent variable $Z$ generates two observation views, $X_S$ and $X_E$, through different transformations.}
    \vspace{-0.3cm}
    \label{fig:graphical model for cca}
\end{figure}

Due to the permutation indeterminacy inherent in ICA \cite{cardoso1993blind}, the Infomax loss can identify independent components but cannot determine their order, resulting in output components that may be misaligned, which is detrimental to effective alignment. To solve this issue, we initialize the two mappings as identity matrices and impose orthogonality constraints on the weights:
\begin{equation}
\label{eq:cross_domain_reg}
\mathcal{L}_{\text{Reg}} = \|\mathbf{H}_{E} \mathbf{H}_{S}^{\top} - \mathbf{I}\|_F,
\end{equation}
where \(\|A\|_F = \sqrt{\sum_{i,j} |a_{ij}|^2}\) represents the Frobenius norm.

Consequently, our loss function is expressed as:

\begin{equation}
\label{eq:main_loss}
\mathcal{L} = \mathcal{L}_{\text{IM}}(X_{S}) +\mathcal{L}_{\text{IM}}(X_{E}) + \lambda \mathcal{L}_{\text{Reg}},
\end{equation}
where $\lambda$ is the hyper-parameter that balances the Infomax loss and the regularization term. In our approach, we treat the synthetic and exemplar domains equally, without involving another balancing coefficient. Visualization of features after disentanglement can be found in \textbf{\cref{fig:random_dim_vis_vertical} in Appendix \ref{sec:more_discussions}}.

\textbf{Inference} 
During inference, given images from both domains, we process them through the entire A\&D pipeline, encoding each image into its feature representation. We then calculate the cosine similarity between feature vectors across domains. The pairs with the highest cosine similarity are considered the most likely matches, providing the final attribution result.

\subsection{Theoretical Analysis}
Given our framework, one may ask why it works without paired data. In this section, we provide a theoretical analysis from a Canonical Correlation Analysis (CCA) perspective, explaining how and why our approach achieves CCA without paired data.

\textbf{CCA Framework.} The data generation process for CCA is illustrated in Figure \ref{fig:graphical model for cca}, in which $Z$ is a latent variable and $\mathbf{X}_{S}$, $\mathbf{X}_{E}$ are two views generated from $Z$ by different transformations. Then, the 
$\mathbf{X}_S \in \mathbb{R}^{m_1}, \mathbf{X}_E \in \mathbb{R}^{m_2}$
denote random variables with covariances \(\mathbf{\Sigma}_{11}, \mathbf{\Sigma}_{22}\) and cross-covariance matrix \(\mathbf{\Sigma}_{12}\). CCA is concerned with finding a pair of linear transformations such that one component within each set of transformed variables is correlated with a single component in the other view \cite{bach2005probabilistic}. 

Then the target for CCA is to find \((\mathbf{H}_{S}^\top \mathbf{X}_S, \mathbf{H}_{E}^\top \mathbf{X}_E)\) that are maximally correlated:
\[
(\mathbf{H}_{S}^*, \mathbf{H}_{E}^*) = \arg\max_{\mathbf{H}_{S}, \mathbf{H}_{E}} \text{corr}(\mathbf{H}_{S}^\top \mathbf{X}_S, \mathbf{H}_{E}^\top \mathbf{X}_E)
\]
\begin{equation}
    = \arg\max_{\mathbf{H}_{S}, \mathbf{H}_{E}} \frac{\mathbf{H}_{S}^\top \mathbf{\Sigma}_{12} \mathbf{H}_{E}}{\sqrt{\mathbf{H}_{S}^\top \mathbf{\Sigma}_{11} \mathbf{H}_{S} \cdot \mathbf{H}_{E}^\top \mathbf{\Sigma}_{22} \mathbf{H}_{E}}}.
    \label{eq:cca}
\end{equation}

Equation \ref{eq:cca} can be equivalently rewritten in the Lagrangian form as:

\begin{equation}
\label{eq:lagrange_cca}
\resizebox{\columnwidth}{!}{$
L =
\underbrace{\mathbf H_S^\top \mathbf{\Sigma}_{12}\mathbf H_E}_{\text{(a) cross-domain correlation}}
-\underbrace{\frac{\lambda}{2}(\mathbf H_S^\top \Sigma_{11}\mathbf H_S-\mathbf{I})
-\frac{\lambda}{2}(\mathbf H_E^\top \mathbf{\Sigma}_{22}\mathbf H_E-\mathbf{I})}_{\text{(b) within-domain uncorrelation}}
$}
\end{equation}

where \(\mathbf{\Sigma}_{12} = \Phi(\mathbf{X}_S)\Phi(\mathbf{X}_E)^\top\), 
\(\mathbf{\Sigma}_{11} = \Phi(\mathbf{X}_S)\Phi(\mathbf{X}_S)^\top\), 
\(\mathbf{\Sigma}_{22} = \Phi(\mathbf{X}_E)\Phi(\mathbf{X}_E)^\top\), and \(\mathbf{H}_S, \mathbf{H}_E \in \mathbb{R}^{d \times d}\) are linear mappings.
Equation ~\eqref{eq:lagrange_cca} decomposes CCA into two goals: (a) keep the two views maximally aligned through a large
cross-domain correlation term $\operatorname{Tr}(\mathbf{H}_S^\top \mathbf{\Sigma}_{12} \mathbf{H}_E)$, and (b) enforce within-domain decorrelation/whitening.
Our method mirrors this decomposition with a two-stage procedure that does not require paired samples.
\paragraph{Stage I: find a shared feature space (alignment).}
We first map both domains into a common representation space using a shared pretrained encoder $\Phi$, producing features
$\Phi(\mathbf{X}_S)$ and $\Phi(\mathbf{X}_E)$ and the empirical cross-covariance $\mathbf{\Sigma}_{12}=\Phi(\mathbf{X}_S) \Phi(\mathbf{X}_E)^\top$. Although we never observe paired $(\mathbf{X}_S,\mathbf{X}_E)$, we \textbf{hypothesize} that contrastive/self-supervised pretraining induces a common representation where the principal semantic axes are approximately aligned between the synthetic and exemplar domains. This is empirically supported by Table \ref{tab:main_result_table} where pretrained backbones alone already yields alignments to some extent.
As a result, $\mathbf{\Sigma}_{12}$ already concentrates much of its mass on “matched” directions, yielding a rough CCA solution to~\eqref{eq:lagrange_cca}.

\paragraph{Stage II: within-domain uncorrelation while preserving cross-domain alignment. (disentanglement)}

We refine each domain with ICA (Infomax) to promote independence, which is a stricter requirement than the uncorrelation constraints in CCA. With the constraint \cref{eq:cross_domain_reg}, we enforce $\mathbf{H}_E \mathbf{H}_S^\top = \mathbf{I}$. Using the cyclic property of the trace operator, we have: $\mathrm{Tr}\left(\mathbf{H}_S^\top \mathbf{\Sigma}_{12} \mathbf{H}_E\right) = \mathrm{Tr}\left(\mathbf{\Sigma}_{12} \mathbf{H}_E \mathbf{H}_S^\top\right) = \mathrm{Tr}(\mathbf{\Sigma}_{12})$ which means that we can optimize the objective in \cref{eq:lagrange_cca}(b) while keeping \cref{eq:lagrange_cca}(a) unchanged, yielding an improved CCA solution. Intuitively, ICA is used to ``clean up'' each view internally (better separated components), while the
identity-biased orthogonal updates prevent the well-aligned cross-domain directions encoded in $\mathbf{\Sigma}_{12}$ from being
destroyed by ICA’s permutation/rotation ambiguity. 
Empirically, we found that replacing \cref{eq:cross_domain_reg} with $\frac{1}{2}\left(\|\mathbf{H}_{E} \mathbf{H}_{E}^{\top} - \mathbf{I}\|_F + \|\mathbf{H}_{S} \mathbf{H}_{S}^{\top} - \mathbf{I}\|_F\right)$ yields better results, thus we conduct experiments with this constraint. \cref{eq:cross_domain_reg} is also effective; results are in \textbf{Appendix~\ref{sec:regularization_details}}.

\section{Experiments}

\begin{figure}[t]
    \centering
    \includegraphics[width=\linewidth]{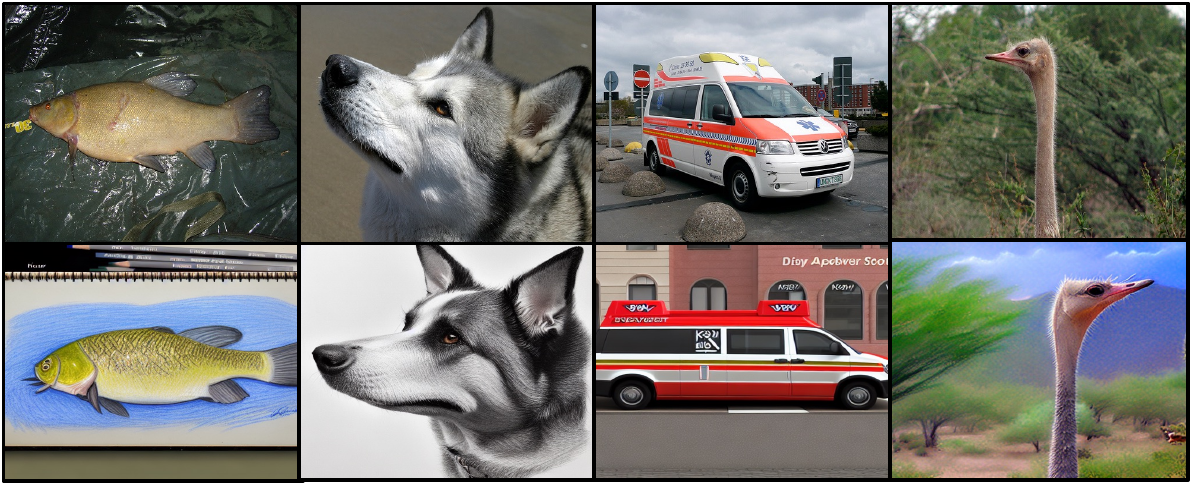}
    \caption{\textbf{Examples of Exemplar (top) and Synthetic (bottom) images in the AbC benchmark.} }
    \vspace{-0.5cm}
    \label{fig:examples}
\end{figure}

\begin{figure*}[h]
    \centering
    \includegraphics[width=0.88\linewidth]{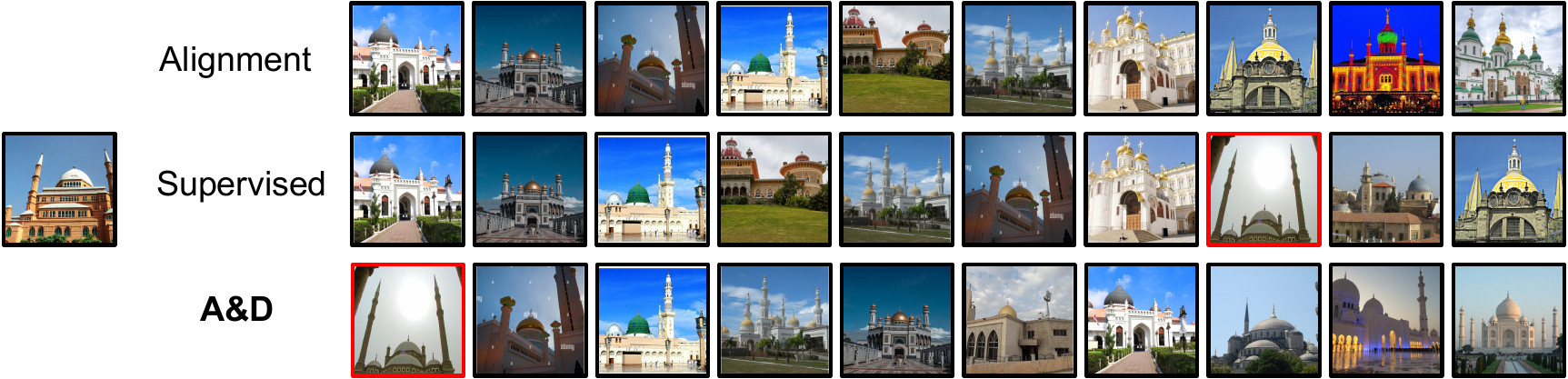}
    \caption{\textbf{Attribution results visualization.} We start with a synthetic image generated by CustomDiffusion (based on Stable Diffusion) and retrieve its exemplars from a pool of 1 million LAION images, showing the top 10 results. \textit{Alignment} denotes attribution using the pretrained DINO model, \textit{Supervised} refers to DINO fine-tuned with paired training data from the AbC benchmark, and A\&D is our proposed unsupervised fine-tuning method. Red bounding boxes indicate the ground-truth exemplars. }
    \vspace{-0.4cm}
    \label{fig:retrived_examples}
\end{figure*}

\subsection{Datasets and Evaluation Metrics}
In the experiments, we apply the AbC (Attribution by Customization) benchmark \cite{wang2023evaluating}, which is specifically designed to evaluate synthetic image attribution for generative models, utilizing Custom Diffusion \cite{kumari2023multi} to generate over 4 million synthesized images influenced by particular exemplar images or artistic styles. It encompasses both object-centric and artistic style synthesis-exemplar image pairs, created using a variety of prompts. Some examples are shown in Figure~\ref{fig:examples}, and more can be found in \textbf{Appendix \ref{sec:more_discussions}}. The benchmark serves as a valuable resource for testing and enhancing attribution algorithms.

The training set consists of three distinct categories: object-centric, artistic styles, and a combination of object-centric and artistic styles. The object-centric images include various object categories sourced from ImageNet \cite{geirhos2018imagenet}, while the artistic style images are drawn from the BAM-FG (Behance Artistic Media - FineGrained) dataset \cite{ruta2021aladin}. The test set is divided into four splits: ImageNet-Seen, BAM-FG, ImageNet-Unseen, and Artchive \cite{Harden_Artchive}. The ImageNet-Seen split contains object categories that are present in the training set, and the BAM-FG split includes artistic styles that are the same as the training data. These two splits serve as in-distribution test cases, closely reflecting the training set’s content. In contrast, the ImageNet-Unseen split comprises object categories absent from the training set, and the Artchive split features artistic styles distinct from those encountered during training. These splits are designed to evaluate the model’s generalization to out-of-distribution data. Additionally, “GPT,” “Media,” and “Object” denote different methods of prompting used to generate the synthetic images. 

Our evaluation setting follows the approach in \cite{wang2023evaluating}. For each synthetic image in the test set, we retrieve its exemplar images from a dataset comprising the exemplars and 1 million images from the LAION dataset \cite{schuhmann2022laion}. The metrics used include Recall@K, which measures the proportion of exemplar images among the top-K retrieved results, and mAP, which evaluates the overall ordering of the retrieval results. Further details about these metrics can be found in the \textbf{Appendix \ref{sec:evaluation_metric_details}}.

\begin{table*}[h]
\setlength{\tabcolsep}{0.5mm}
\centering
\caption{\textbf{R@5 and mAP of four scenarios on AbC benchmark.} We compare A\&D with both supervised method and unsupervised methods (Pretrain, Pseudo label~\cite{lee2013pseudo}) on five different backbones across four subdatasets. We empirically find that A\&D outperforms the unsupervised baselines in all settings except for Pseudo Labeling using the SSCD backbone and, surprisingly, surpasses the supervised method when using the MOCO, DINO, and ViT as backbones.}
\resizebox{\textwidth}{!}{%
\begin{tabular}{llcccccccccccccccc|cc}
\toprule
\multicolumn{2}{l}{Source} & \multicolumn{4}{c}{ImageNet-Seen} & \multicolumn{4}{c}{BAM-FG} & \multicolumn{4}{c}{ImageNet-Unseen} & \multicolumn{4}{c|}{Artchive} 
& 

\\ \cmidrule(lr){1-2} \cmidrule(lr){3-6} \cmidrule(lr){7-10} \cmidrule(lr){11-14} \cmidrule(lr){15-18} \cmidrule(lr){19-20}
\multicolumn{2}{l}{Prompts} & \multicolumn{2}{c}{GPT} & \multicolumn{2}{c}{Media} & \multicolumn{2}{c}{GPT} & \multicolumn{2}{c}{Object} & \multicolumn{2}{c}{GPT} & \multicolumn{2}{c}{Media} & \multicolumn{2}{c}{GPT} & \multicolumn{2}{c|}{Object} & 

\multicolumn{2}{|c}{\textbf{Average}}
\\ \cmidrule(lr){1-2} \cmidrule(lr){3-4} \cmidrule(lr){5-6} \cmidrule(lr){7-8} \cmidrule(lr){9-10} \cmidrule(lr){11-12} \cmidrule(lr){13-14} \cmidrule(lr){15-16} \cmidrule(lr){17-18}

$F_\text{base}$ & Method & R@5 & mAP & R@5 & mAP & R@5 & mAP & R@5 & mAP & R@5 & mAP & R@5 & mAP & R@5 & mAP & R@5 & mAP & R@5 & mAP\\ \midrule

\multirow{5}{*}{MoCo}  

&Pretrain & 0.390 & 0.347 & 0.239 & 0.211    & 0.130 & 0.144 & 0.187 & 0.211   & 0.761 & 0.717 & 0.460 & 0.408   & 0.169 & 0.164 & 0.131 & 0.125 & 0.308 & 0.291 \\
&Pseudo & 0.418 & 0.377 & 0.251 & 0.221 & 0.132 & 0.146 & 0.187 & 0.212 & 0.778 & 0.736 & 0.452 & 0.402 & 0.166 & 0.162 & 0.132 & 0.128 & 0.315 & 0.298 \\
&A\&D  & 0.444 & 0.399 & 0.313 & 0.277 & 0.157 & 0.175 & 0.209 & 0.238 & 0.804 & 0.765 & 0.569 & 0.512 & 0.204 & 0.204 & 0.158 & 0.154 & \textbf{0.357} & \textbf{0.341} \\

\rowcolor{gray!40}&Superv & 0.437 & 0.394 & 0.295 & 0.262   & 0.153 & 0.172 & 0.209 & 0.238   & 0.791 & 0.753 & 0.519 & 0.467   & 0.208 & 0.209 & 0.165 & 0.165 & 0.347 & 0.333 \\

\midrule

\multirow{5}{*}{DINO}  

&Pretrain & 0.433 & 0.393 & 0.288 & 0.255    & 0.148 & 0.163 & 0.193 & 0.219   & 0.831 & 0.795 & 0.540 & 0.492   & 0.183 & 0.181 & 0.140 & 0.136  & 0.345 & 0.329 \\

&Pseudo & 0.471 & 0.426 & 0.374 & 0.330 & 0.127 & 0.142 & 0.164 & 0.187 & 0.840 & 0.804 & 0.639 & 0.588 & 0.211 & 0.216 & 0.161 & 0.160  & 0.373 & 0.357 \\
&A\&D  & 0.491 & 0.448 & 0.364 & 0.323 & 0.167 & 0.186 & 0.210 & 0.239 & 0.845 & 0.811 & 0.633 & 0.581 & 0.212 & 0.214 & 0.164 & 0.161 & \textbf{0.386} & \textbf{0.370} \\

\rowcolor{gray!40}
&Superv & 0.475 & 0.433 & 0.351 & 0.311   & 0.165 & 0.184 & 0.212 & 0.239   & 0.842 & 0.810 & 0.598 & 0.549 & 0.217 & 0.221 & 0.170 & 0.169 & 0.379 & 0.365  \\

\midrule

\multirow{5}{*}{ViT} 
& Pretrain   & 0.355 & 0.310 & 0.242 & 0.210    & 0.168 & 0.193 & 0.224 & 0.259    & 0.785 & 0.726 & 0.474 & 0.410 & 0.201  & 0.210 & 0.156 & 0.161 & 0.326 & 0.310\\

& Pseudo     & 0.361 & 0.319 & 0.266 & 0.230 & 0.103 & 0.117 & 0.141 & 0.162 & 0.794 & 0.737 & 0.525 & 0.457 & 0.156 & 0.161 & 0.115 & 0.114 & 0.308 & 0.287 \\
& A\&D   & 0.440 & 0.395 & 0.295 & 0.259 & 0.203 & 0.229 & 0.251 & 0.288 & 0.845 & 0.805 & 0.561 & 0.500 & 0.229 & 0.240 & 0.176 & 0.182 & \textbf{0.375} & \textbf{0.362} \\

\rowcolor{gray!40}
& Superv  & 0.448 & 0.398 & 0.315 & 0.274    & 0.182 & 0.209 & 0.234 & 0.272    & 0.820 & 0.772 & 0.539 & 0.474 & 0.206 & 0.216 & 0.159 & 0.163 & 0.363 & 0.347\\

\midrule

\multirow{5}{*}{CLIP}
& Pretrain   & 0.236 & 0.195 & 0.137 & 0.118 & 0.129 & 0.148 & 0.174 & 0.200   & 0.580 & 0.500 & 0.239 & 0.201 & 0.186 & 0.198 & 0.140 & 0.144 & 0.228 & 0.213 \\

& Pseudo     & 0.018 & 0.015 & 0.014 & 0.011 & 0.022 & 0.026 & 0.032 & 0.036 & 0.032 & 0.027 & 0.009 & 0.007 & 0.185 & 0.212 & 0.147 & 0.161 & 0.057 & 0.062 \\
& A\&D   & 0.257 & 0.212 & 0.153 & 0.132 & 0.173 & 0.202 & 0.216 & 0.251 & 0.607 & 0.525 & 0.258 & 0.218 & 0.229 & 0.256 & 0.176 & 0.190 & 0.259 & 0.248 \\

\rowcolor{gray!40}
& Superv   & 0.329 & 0.280 & 0.222 & 0.190 & 0.186 & 0.219 & 0.236 & 0.278   & 0.701  & 0.633 & 0.389  & 0.332 & 0.259  & 0.297 & 0.201 & 0.223 & \textbf{0.315} & \textbf{0.306} \\

\midrule

\multirow{5}{*}{SSCD}  

&Pretrain & 0.253 & 0.230 & 0.142 & 0.131   & 0.117 & 0.128 & 0.175 & 0.194   & 0.601 & 0.566 & 0.264 & 0.245    & 0.151 & 0.142 & 0.123 & 0.115 & 0.228 & 0.219 \\

&Pseudo & 0.267 & 0.241 & 0.154 & 0.140 & 0.110 & 0.118 & 0.161 & 0.178 & 0.618 & 0.581 & 0.285 & 0.263 & 0.156 & 0.147 & 0.126 & 0.119 & 0.235 & 0.223 \\
&A\&D & 0.259 & 0.234 & 0.147 & 0.134 & 0.119 & 0.130 & 0.176 & 0.196 & 0.610 & 0.573 & 0.272 & 0.251 & 0.150 & 0.142 & 0.123 & 0.115 & 0.232 & 0.222 \\

\rowcolor{gray!40}
&Superv & 0.268 & 0.242 & 0.156 & 0.142  & 0.118 & 0.128 & 0.314 & 0.193   & 0.621 & 0.586 & 0.282 & 0.261   & 0.159 & 0.150 & 0.131  & 0.123  & \textbf{0.256} & \textbf{0.228} \\

\bottomrule
\end{tabular}%
}
\vspace{-0.3cm}
\label{tab:main_result_table}
\end{table*}

\subsection{Implementation details}

For alignment, we experimented with different pretrained models, including contrastive self-supervised approaches (MoCo V3 \cite{chen2021empirical} and DINO \cite{caron2021emerging}), a supervised method (ViT \cite{dosovitskiy2020image}), and a copy detection model (SSCD \cite{pizzi2022self}). To ensure fair comparison, we used the same ViT-B/16 backbone architecture across MoCo, DINO, and ViT. For disentanglement, we fine-tuned two separate square linear mappings for the two domains using the Adam optimizer \cite{kingma2014adam}, with a learning rate of 0.001. We set two hyperparameters: \(\lambda\) controls \(\mathcal{L}_{Reg}\) and \(\mu\) controls \(\mathcal{L}_{IM}\). The values of \(\lambda\) were set to 0.1 for MoCo, DINO, and CLIP, and 0.15 for ViT and SSCD, while \(\mu\) was set to 1e3 for MoCo and 1 for all others. Our models were fine-tuned on a single A100 40G GPU with a batch size of 512 for ViT and 1024 for the other models. All the experiments are implemented by Pytorch 2.3.0.

\subsection{Comparison with other methods}
We conducted a comprehensive evaluation of our proposed A\&D method against several other approaches, including both supervised and unsupervised methods: Supervised Learning \cite{wang2023evaluating}, Pseudo Labeling \cite{lee2013pseudo}, and Pretraining (inference without further fine-tuning). The models were tuned on three training sets from the AbC benchmark: (1) object-centric, (2) artistic styles, and (3) a combination of both. We evaluated the performance of models trained on the combined training set (3) across all test sets, with results presented in terms of Recall@5 and mAP, as shown in Table \ref{tab:main_result_table}. 

Our method, A\&D, a fully unsupervised approach, outperformed all other methods, including supervised ones, in the average of both Recall@5 and mAP across MoCo, DINO, and ViT backbones. Notably, A\&D achieved the highest average Recall@5 (0.386) and mAP (0.370) with the DINO backbone. Compared to other unsupervised methods, our approach outperformed all except Pseudo Labeling on the SSCD backbone, where it was only 0.3\% and 0.1\% lower in Recall@5 and mAP, respectively. On the ImageNet-Unseen Media test set, our A\&D method, using MoCo as the alignment backbone, outperformed Pretraining, Pseudo Labeling, and Supervised methods by 10.9\%, 11.7\%, and 5\% in Recall@5, and by 10.4\%, 11\%, and 4.5\% in mAP, respectively. These results underscore the effectiveness of our unsupervised method for synthetic image attribution. Additional experimental results on the AbC benchmark can be found in \textbf{Appendix \ref{sec:full_experimental_results}}, and visualization of attribution results can be found in \textbf{Appendix \ref{sec:visualization_of_attribution}}.

We also visualize the retrieved results in \cref{fig:retrived_examples}. 
Compared with using the alignment backbone alone (Alignment) or supervised fine-tuning (Supervised), our A\&D tends to produce more fine-grained concept matching.
In the example shown, Alignment and Supervised mainly retrieve images that share the coarse concept (``mosque'') but vary substantially in viewpoint and architectural details, whereas A\&D ranks exemplars that better preserve the specific visual cues of the query—e.g., a similar frontal composition with the two tall minarets framing the central dome and sharper, peaked silhouettes—thereby yielding a more detailed cross-domain alignment. More examples can be found in \textbf{Appendix \cref{fig:visualization_of_attribution_results}}.

\subsection{Discussion}

\begin{figure*}[h]
    \centering
    \includegraphics[width=0.7\linewidth]{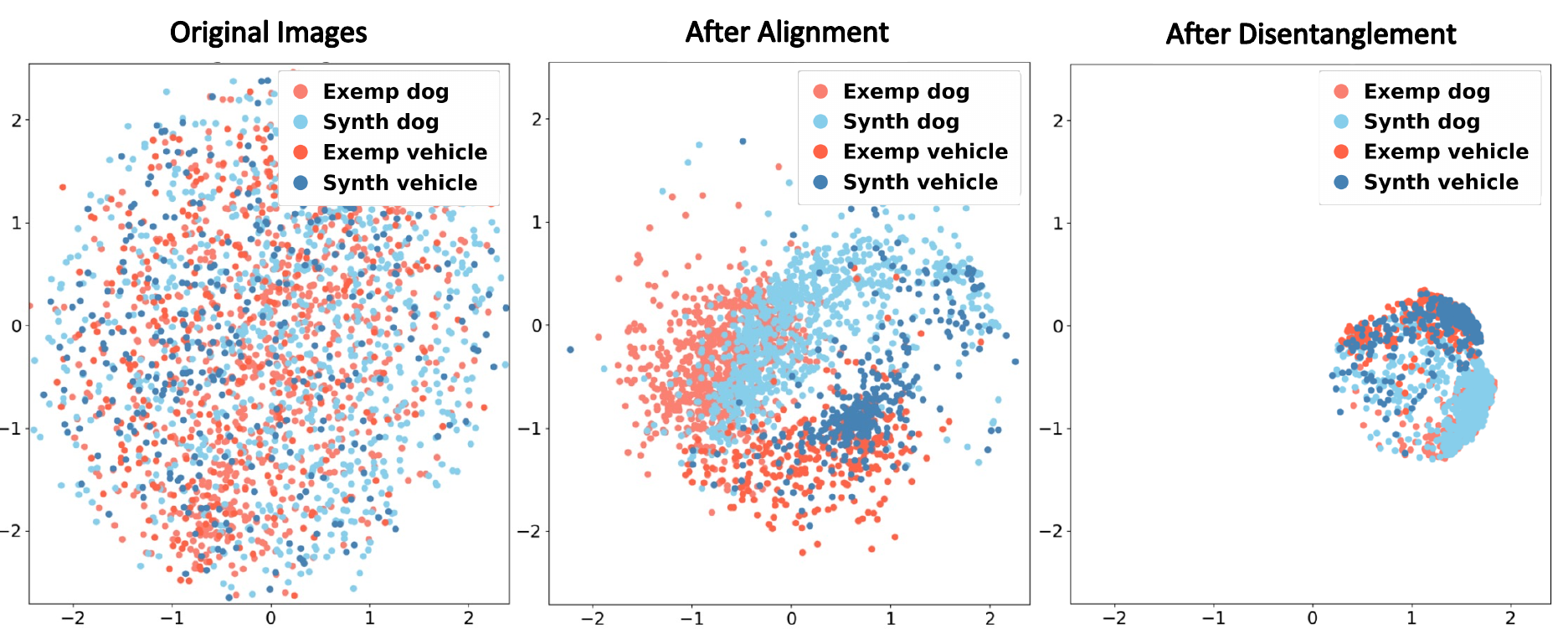}
    \caption{\textbf{Scatter Plot of Distribution Changes through A\&D using MoCo alignment backbone.} The scatter plot illustrates the feature distribution changes for two categories of images from AbC’s object-centric training set: dog and vehicle. The original data points appear almost random, with little to no alignment between domains. After alignment, we observe that corresponding objects from the two views become more closely matched. With disentanglement, this alignment becomes significantly more distinct and precise, demonstrating the effectiveness of our A\&D framework. } 
    \vspace{-0.3cm}
    \label{fig:tsne_features}
\end{figure*}

\paragraph{What's the Difference between Our A\&D and Adaptation?}
We would like to highlight that our approach is not simply an adaptation of pretrained models under distribution shift, where fine-tuning on new data could be sufficient—a direction that has been widely explored in prior work. While adaptation typically involves a single domain undergoing distribution shift, our A\&D framework considers two distinct domains, making it conceptually different. In contrast, our A\&D framework operates across two distinct domains. To empirically validate this distinction, we fine-tuned DINO on the BAM-FG exemplar training set for 10 epochs and used the resulting model as the alignment backbone in our disentanglement process. As shown in Table \ref{tab:DINO-tuned}, fine-tuning DINO on BAM-FG yields only marginal improvements, with less than a 1\% increase in both Recall@5 and mAP on the BAM-FG test set. Furthermore, comparing the two DINO models as alignment backbones reveals that fine-tuning on exemplars does not meaningfully enhance A\&D performance, as Recall@5 and mAP remain nearly unchanged with or without fine-tuning.
\vspace{-0.2cm}
\paragraph{Why Supervised Methods Underperform Unsupervised?}
The supervised method underperforms due to the ambiguous nature of the supervision. Such noise may introduce misleading signals, resulting in suboptimal performance. An example of the dataset is shown in \cref{fig:noisy-dataset-visualization}.

\begin{table}[t]
\setlength{\tabcolsep}{1mm}
\centering
\caption{\textbf{Performance of DINO tuned on BAM-FG.} ``Pretrain" refers to the original DINO model, ``Tuned" indicates the DINO model further fine-tuned for 10 epochs on BAM-FG, and ``Tuned+A\&D" denotes the model trained on the object + style training set of AbC.}
\resizebox{0.35\textwidth}{!}{%
\begin{tabular}{llcccc}
\toprule
\multicolumn{2}{l}{Source} & \multicolumn{4}{c}{BAM-FG} \\
\cmidrule(lr){1-2} \cmidrule(lr){3-6}
\multicolumn{2}{l}{Prompts} & \multicolumn{2}{c}{GPT} & \multicolumn{2}{c}{Object} \\
$F_\text{base}$ & Method & R@5 & mAP & R@5 & mAP  \\ \midrule

\multirow{4}{*}{DINO}
& Pretrain  & 0.148 & 0.163 & 0.193 & 0.219 \\
& Tuned  & 0.154 & 0.172 & 0.198 & 0.227  \\
& A\&D  & \textbf{0.167} & \textbf{0.186} & \textbf{0.210} & 0.239 \\
& Tuned+A\&D & 0.162 & 0.184 & 0.209 & \textbf{0.243} \\

\bottomrule
\end{tabular}
}
\vspace{-0.3cm}
\label{tab:DINO-tuned}
\end{table}

\begin{figure}[t]
    \centering
    \includegraphics[width=\linewidth]{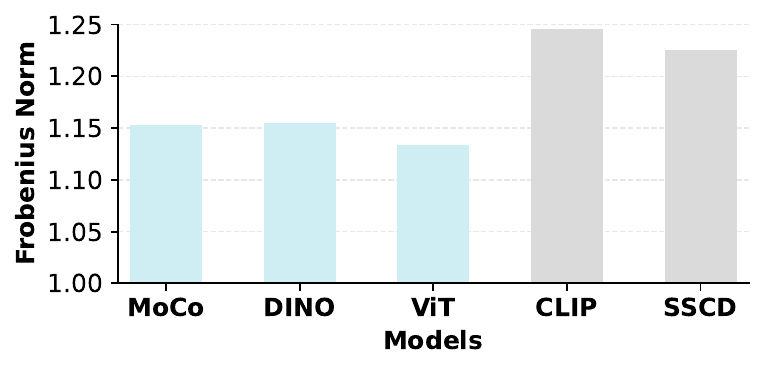}
    \caption{\textbf{F-norms for Different Backbone Models.} We show the F-norm of the difference between the covariance matrix of synthetic and exemplar features from AbC’s object-centric training set and identity matrix, compared across different pretrained backbone models. The lower F-norms observed for MoCo, DINO, and ViT compared to CLIP and SSCD indicate better alignment. }
    \label{fig:Frobenius_Norms_for_Different_Models}
    \vspace{-0.3cm}
\end{figure}

\paragraph{Why the Moderate Performance?}
The moderate performance of A\&D with CLIP and SSCD can be attributed to the inherent characteristics of these models. CLIP, trained using two different modalities—text and image—might not produce representations that are well-suited for cross-domain image alignment. The pseudo-labeling results for CLIP in Table \ref{tab:main_result_table} reveal that using pretrained CLIP actually misdirects the training process, leading to suboptimal alignment. This observation is further supported by the Frobenius norm analysis in Figure \ref{fig:Frobenius_Norms_for_Different_Models}. For SSCD, which is a contrastive self-supervised backbone specifically trained for copy detection, our A\&D framework underperforms compared to even pseudo-labeling. We believe this is because SSCD’s learned features are specialized for copy detection, lacking the generalization needed to effectively represent more complex image distributions across domains.

\paragraph{Distribution Visualization} Figure \ref{fig:tsne_features} demonstrates the efficacy of our proposed framework in aligning features across domains. We selected two object categories (vehicle and dog) from AbC’s object-centric training set and visualized their feature distribution changes. In these scatter plots, red points correspond to images from the exemplar domain, while blue points represent images from the synthetic domain. Point darkness encodes category shared across domains: darker points denote one category, lighter points the other. Initially, the distributions of the original images are scattered randomly, showing significant overlap and poor cross-domain alignment. After passing the alignment backbone which we used MoCo here, while not perfect, there is a noticeable grouping of the same category across domains, indicating that we have successfully started to align the features. After disentanglement, the distributions tighten and separate more clearly by category, indicating improved cross-domain alignment. This refinement highlights disentanglement’s role in achieving stronger alignment.

\section{Conclusion}
In this paper, we have proposed an unsupervised framework for synthetic image attribution, termed Alignment and Disentanglement. Our approach eliminates the need for paired annotations by using contrastive self-supervised learning for alignment and ICA for disentanglement. Through theoretical analysis, we demonstrated that our framework aligns with the CCA framework under some intuitive assumptions, offering a new perspective on implementing CCA without paired data. Our extensive experiments on the AbC benchmark show that A\&D not only matches but, in some cases, surpasses the performance of the supervised method. 
\textbf{Limitations:} we assume access to a pretrained backbone with reasonable cross-domain alignment; addressing larger domain gaps is an important direction for future work. One solution is to add an explicit cross-domain alignment stage, trained with self-supervision and stability regularization, to better handle large domain gaps. We hope that this work, including the findings and algorithm, will shed light on more efficient approaches for synthetic image attribution and inspire further research to design more effective methods.

\section*{Impact Statement}

This paper presents work whose goal is to advance the field of 
Machine Learning. There are many potential societal consequences 
of our work, none which we feel must be specifically highlighted here.

\nocite{langley00}

\bibliography{example_paper}
\bibliographystyle{icml2026}

\newpage
\appendix
\onecolumn
\begin{center}
{\LARGE \textbf{Appendix}}
\end{center}

\section{Difference from Model Attribution}
\label{sec:difference_from_model_attribution}

For a synthesized image, recent works have focused on identifying the image’s source model \cite{yu2019attributing, bui2022repmix, marra2019gans, sha2023fake} or determining which model can best represent it \cite{lu2023content}, a field known as model attribution. In the context of GAN-based models, membership inference, an open problem, has been explored for both generative \cite{hayes2017logan, hilprecht2019monte, chen2020gan, carlini2021extracting} and discriminative \cite{shokri2017membership, sablayrolles2018d, hu2022membership} models. The main difference in our work is that we don’t focus on the model, but rather on the synthesized images and their corresponding training data.

\section{Setting Comparison with Domain Generalization}
\label{sec:dg_comparison}
\begin{figure}
    \centering
    \includegraphics[width=0.55\linewidth]{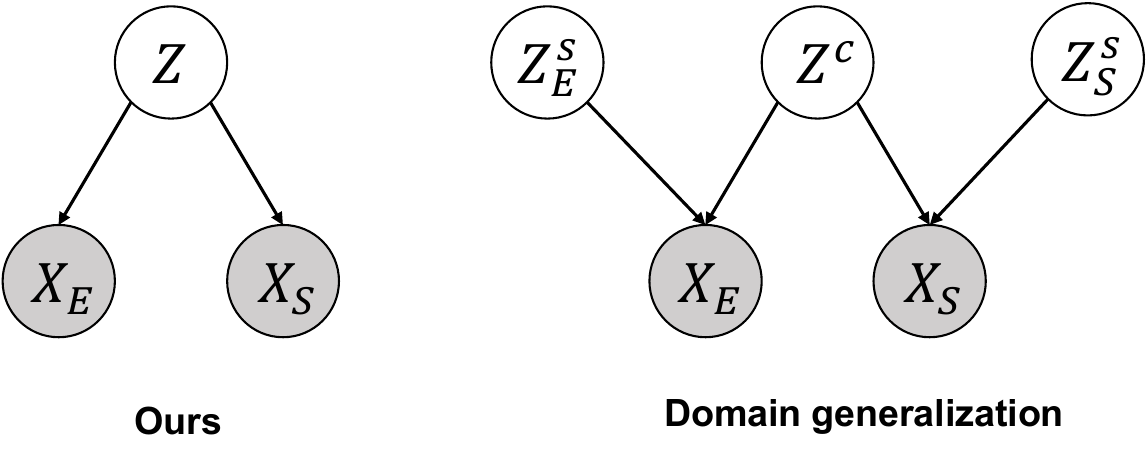}
    \caption{\textbf{Data generation process for comparing the difference between our framework and domain generalization.}}
    \label{fig:difference_with_domain_generalization}
\end{figure}

While our approach and domain generalization share the high-level goal of bridging two domains, the underlying assumptions about the data generation process are fundamentally different, below we show the detailed difference of the data generation process. An illustration of the difference is given in \cref{fig:difference_with_domain_generalization}.

\paragraph{Domain generalization.}
A common assumption in domain generalization is that data from different domains are generated from a shared invariant latent variable $\mathbf{Z}^c$ and domain-specific variables $\mathbf{Z}_E^s$ and $\mathbf{Z}_S^s$. Specifically, the generation follows:
\begin{equation}
\mathbf{X}_E = g(\mathbf{Z}^c, \mathbf{Z}_E^s)
\quad \text{and} \quad
\mathbf{X}_S = g(\mathbf{Z}^c, \mathbf{Z}_S^s),
\end{equation}
where the generation function $g$ is shared. The objective is to recover $\mathbf{Z}^c$---the shared, invariant component---across multiple domains, enabling better generalization to unseen ones.

\paragraph{Our framework.}
In contrast, our framework assumes that both $\mathbf{X}_E$ and $\mathbf{X}_S$ are generated from a single shared latent variable $\mathbf{Z}$, but through distinct functions:
\begin{equation}
\mathbf{X}_E = g_E(\mathbf{Z}, \boldsymbol{\epsilon}_E)
\quad \text{and} \quad
\mathbf{X}_S = g_S(\mathbf{Z}, \boldsymbol{\epsilon}_S),
\end{equation}
where $\boldsymbol{\epsilon}_E$ and $\boldsymbol{\epsilon}_S$ are independent noise terms. Instead of recovering a shared component from partially shared inputs, we aim to learn two distinct functions that map both $\mathbf{X}_E$ and $\mathbf{X}_S$ into a common latent space $\mathbf{Z}$.

\section{Evaluation Metric Details}
\label{sec:evaluation_metric_details}

\textbf{Recall@k} measures the fraction of relevant exemplar images that are successfully retrieved among the top-k results. It is computed as:

\[
\text{Recall@k} = \frac{\text{$n_e$}}{\text{$n_a$}},
\]

where $n_e$ is the number of $X_e^{(i)}$ in top k, $n_a$ is total number of $X_e^{(i)}$.\\\\
\textbf{Mean Average Precision(mAP)} takes into account both the precision and ranking quality of the retrieved results. The Average Precision (AP) for a single synthetic image is calculated as:

\[\text{AP} = \sum_{k=1}^{m} \left(\frac{P(k) \cdot \text{Rel}(k)}{\text{$n_e$}}\right), 
\]

where $P(k)$ is the precision at cut-off $k$ in the list, $\text{Rel}(k)$ is an indicator function equaling 1 if the item at rank $k$ is a matched exemplar image, and 0 otherwise. $m$ is the total number of images retrieved.

The mAP is then computed as the mean of AP scores over all synthetic images used for evaluation:

\[
\text{mAP} = \frac{\sum_{i=1}^{s} \text{AP}_i}{s},
\]

where $\text{AP}_i$ is the average precision score for the $i$-th synthetic image, and $s$ is the total number of synthetic images.

\begin{figure}
    \centering
    \includegraphics[width=0.5\linewidth]{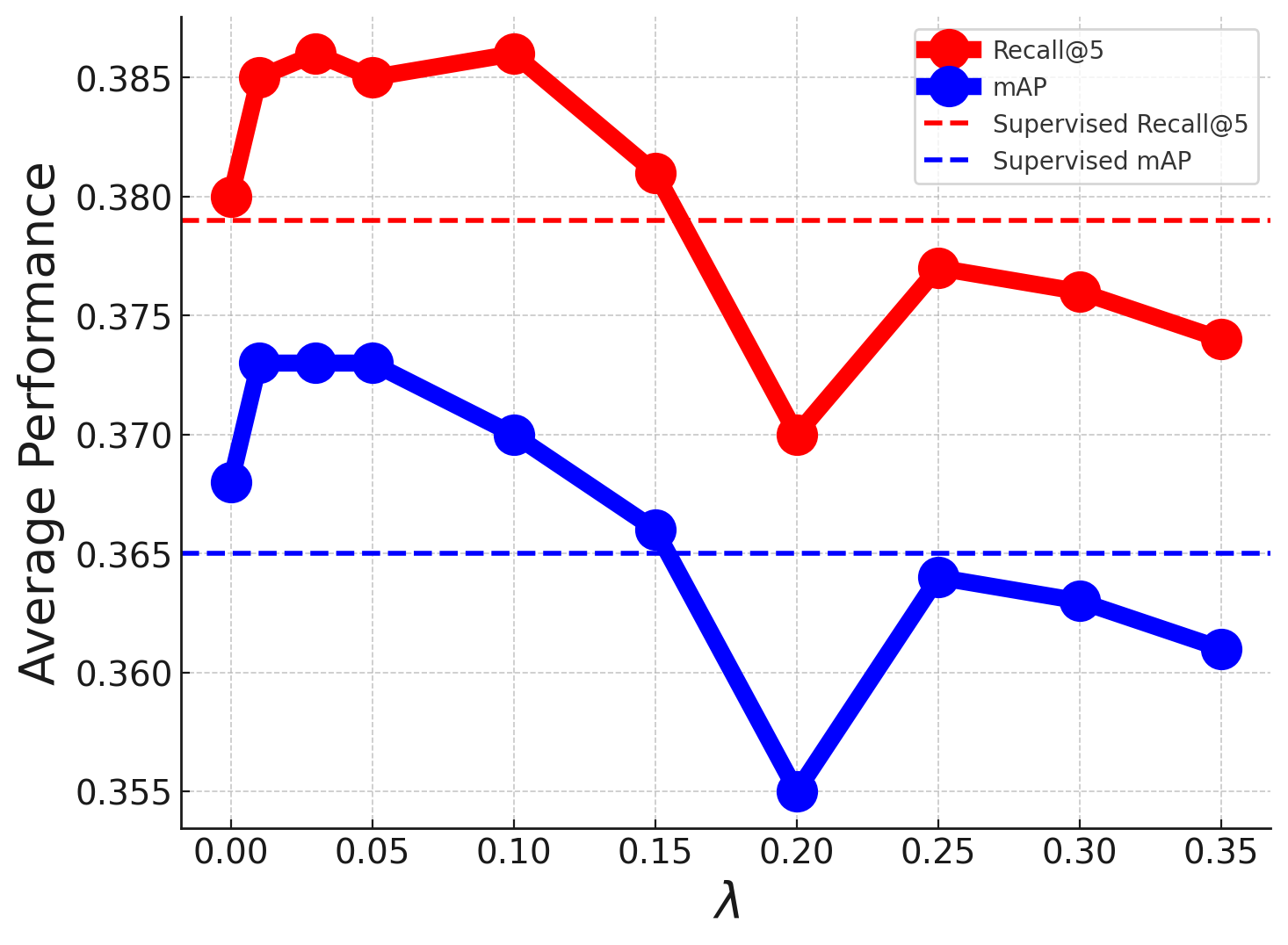}
    \caption{\textbf{Average Recall@5 and mAP for different $\lambda$.} A\&D with the DINO alignment backbone trained on the object+style training set.}
    \label{fig:different_lambda}
\end{figure}

\section{Visualization of attribution results}
\label{sec:visualization_of_attribution}
\begin{figure}
    \centering
    \includegraphics[width=1\linewidth]{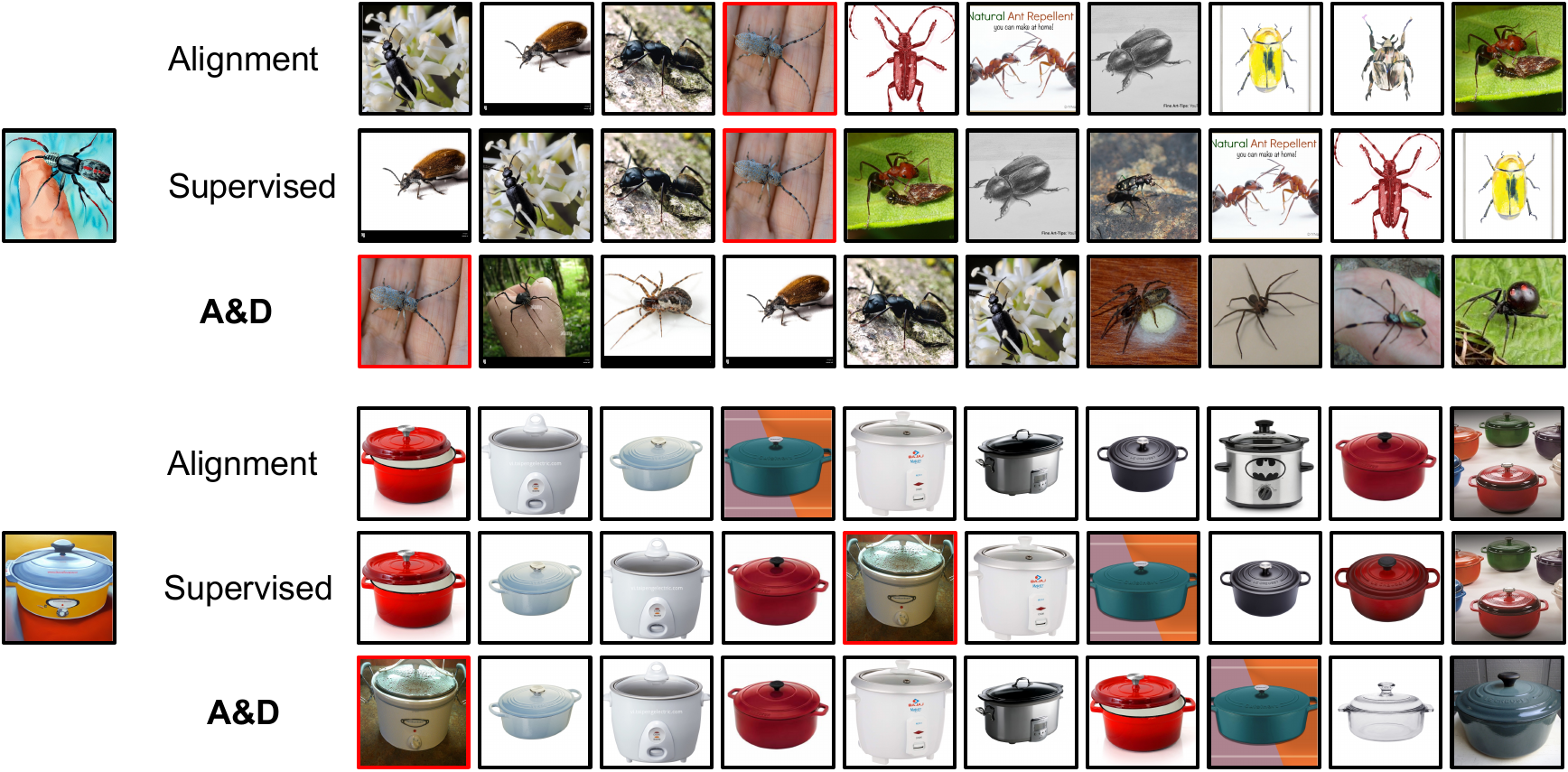}
    \caption{\textbf{More attribution results visualization.}}
    \label{fig:visualization_of_attribution_results}
\end{figure}

As illustrated in Figure \ref{fig:visualization_of_attribution_results}, all methods retrieve content-relevant images rather than merely matching based on color. For instance, in the “\emph{spider on a finger}” example, the original background is blue, yet none of the top-10 matches from any method contain a blue background—highlighting that semantic cues (e.g., the spider) dominate over low-level color similarity. Although all approaches successfully retrieve the correct exemplar, A\&D appears to rank more semantically aligned images higher, such as the second and ninth results showing a spider on a hand. In the “\emph{pot on a red desk}” example, color does influence retrieval to some extent: red pots appear among the top matches. This is reasonable, as color can contribute to similarity. However, A\&D strikes a better balance between semantic content and color, retrieving only two red pots, while the Supervised model retrieves four and even ranks a red pot as the top result.

\section{More Discussions}
\label{sec:more_discussions}
\paragraph{Does Alignment Matter?}
\textbf{Yes!} Figure \ref{fig:Frobenius_Norms_for_Different_Models} presents the Frobenius norms of the difference between the cross-domain covariance matrix and the identity matrix, computed for synthetic and exemplar features from AbC’s object-centric training set. The cross-domain covariance matrix converges towards the identity matrix as the alignment between domains improves. Therefore, a lower Frobenius norm directly correlates with better alignment. Our results reveal that MoCo, DINO, and ViT exhibit significantly lower F-norms, indicating superior alignment compared to CLIP and SSCD. This finding aligns with the performance outcomes of our A\&D method, as detailed in Table \ref{tab:main_result_table}. The choice of alignment backbone is crucial, as stronger backbones enable our A\&D method to potentially surpass even supervised approaches in cross-domain alignment.

\paragraph{Is Contrastive Alone Sufficient?}
\textbf{No.} As demonstrated in Table \ref{tab:main_result_table}, our framework using the ViT alignment backbone ranks third among all 20 cases in terms of average Recall@5 and mAP. This indicates that large-scale supervised pretraining also plays a significant role in enhancing alignment, complementing contrastive learning approaches.

\paragraph{Combing A\&D with other methods}
In Table \ref{tab:AD + pseudo and supervise}, we used DINO as the alignment backbone and combined our A\&D method with both supervised learning and pseudo-labeling approaches. The results indicate that A\&D does not consistently enhance its performance. For instance, while the combination of A\&D with pseudo-labeling shows an average improvement in Recall@5 and mAP, it leads to a performance drop on the ImageNet-Seen test set and the ImageNet-Unseen set when prompted with Media. A similar trend is observed when combining A\&D with the supervised method: although there is an improvement on the ImageNet-Seen and ImageNet-Unseen sets, there is a noticeable degradation in performance on the BAM-FG and Artchive datasets.

\begin{figure}
    \centering
    \includegraphics[width=0.6\linewidth]{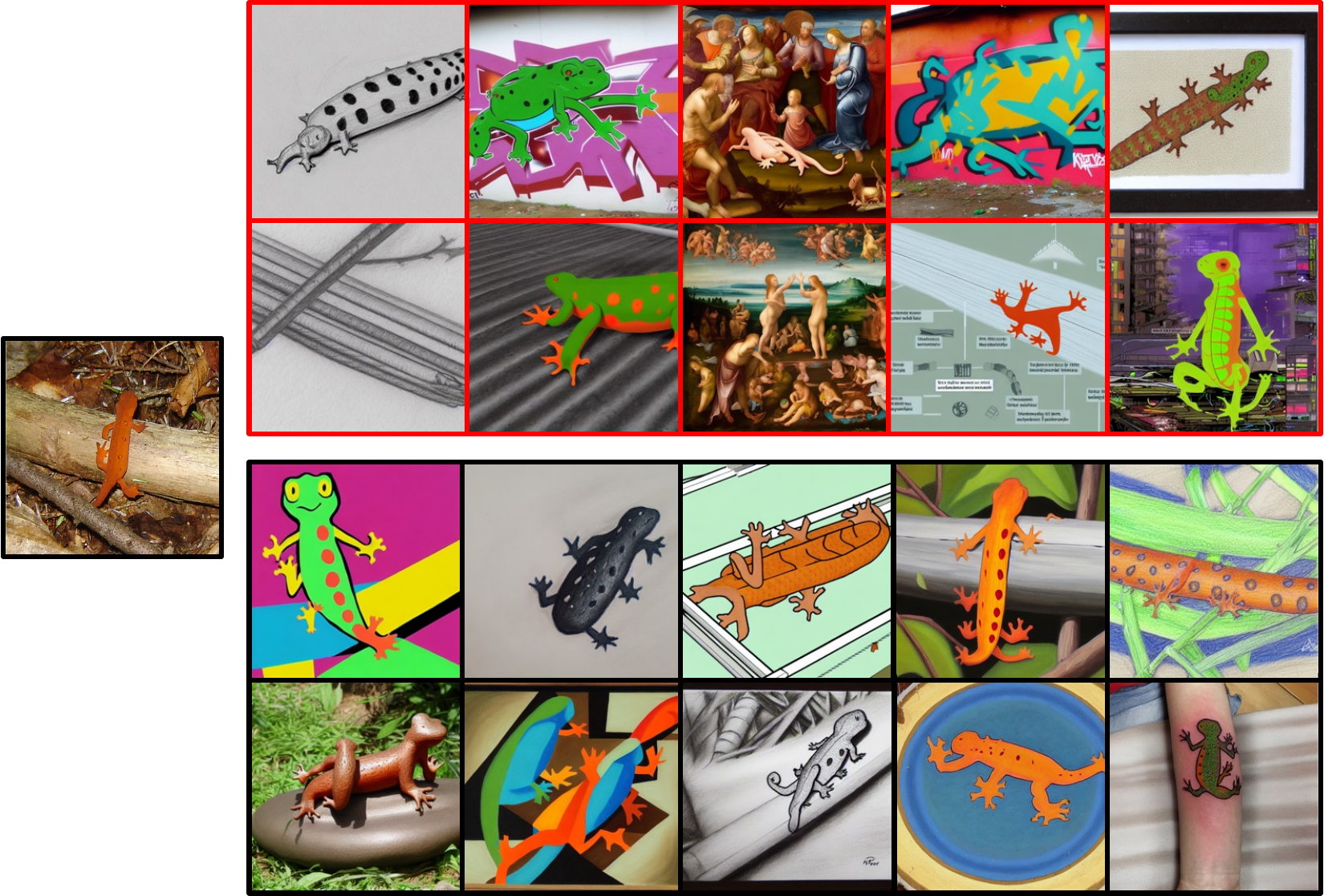}
    \caption{\textbf{Visualization of the object-centric training set in the AbC benchmark.} The image on the left shows a real exemplar, while the 20 images on the right illustrate a subset of its paired synthetic counterparts. In the AbC benchmark, the quality of synthetic data varies significantly. Some images, such as those outlined in black at the bottom, provide strong supervisory signals. In contrast, many others—such as those highlighted in red at the top—are noisy or ambiguous, potentially hindering effective training. These 20 examples represent only a small portion of the hundreds of synthetic images generated per exemplar and do not fully capture the overall variability and noise present in the dataset.}
    \label{fig:noisy-dataset-visualization}
\end{figure}

\paragraph{Random initialization for $\mathbf{H}_{S}$ and $\mathbf{H}_{E}$}
We used identity initialization in our method and also experimented with random initialization. Specifically, we tried initializing the two linear mappings with both the same and different random matrices. In Table \ref{tab:ablation study}, we observe that when the two linear mappings are initialized with the same random matrix, the performance of our A\&D method does not degrade significantly and remains comparable to the supervised method. However, when the mappings are initialized with different random matrices, our A\&D method fails across all alignment backbones on all test sets. This is because identical initialization minimizes the likelihood of permutation occurring during ICA, thereby achieving alignment. Conversely, when initialized with different matrices, permutation almost certainly occurs, making alignment impossible.

\begin{figure}[htbp]
    \centering
    
    \includegraphics[width=0.6\textwidth]{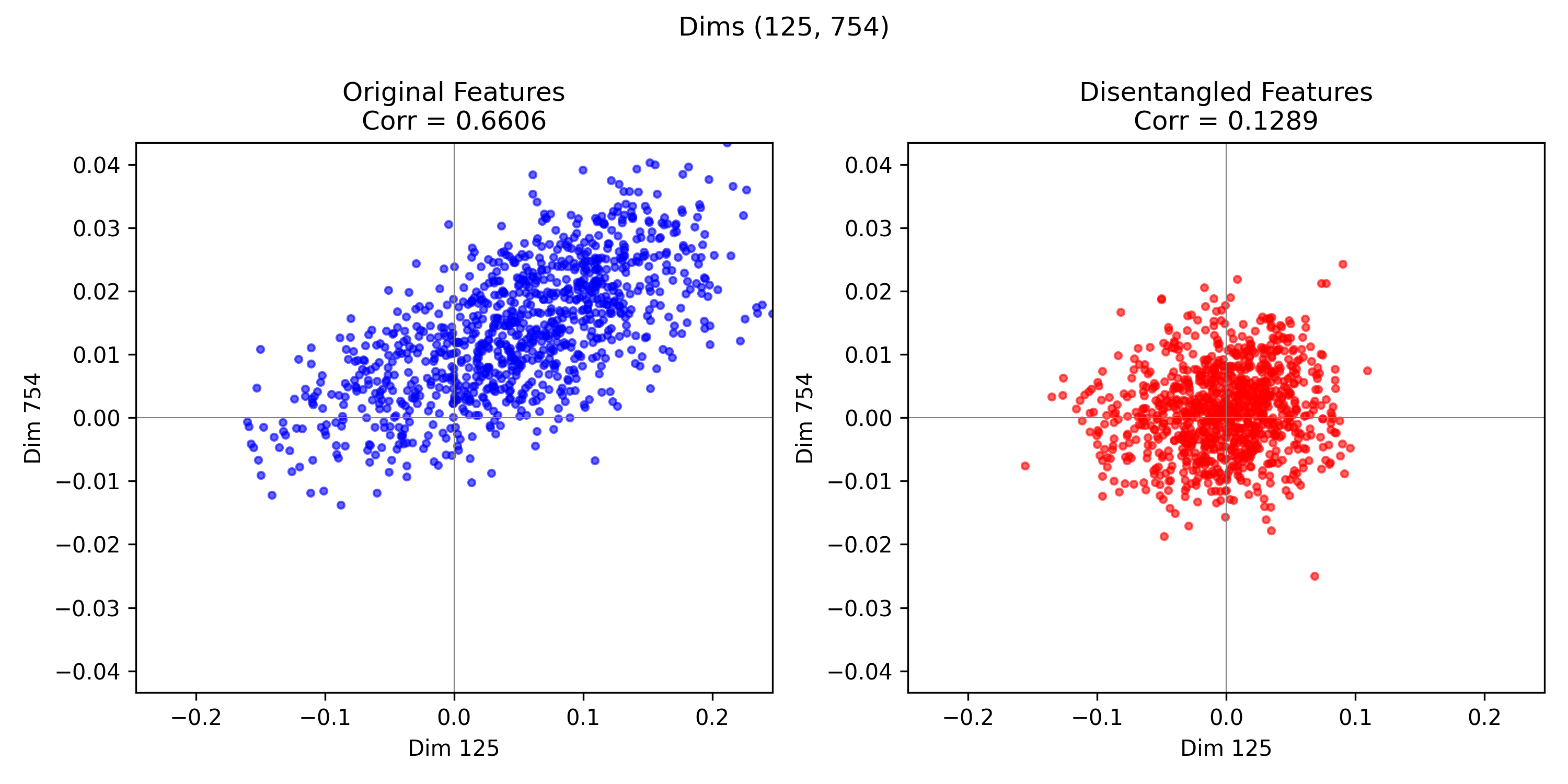}

    
    \includegraphics[width=0.6\textwidth]{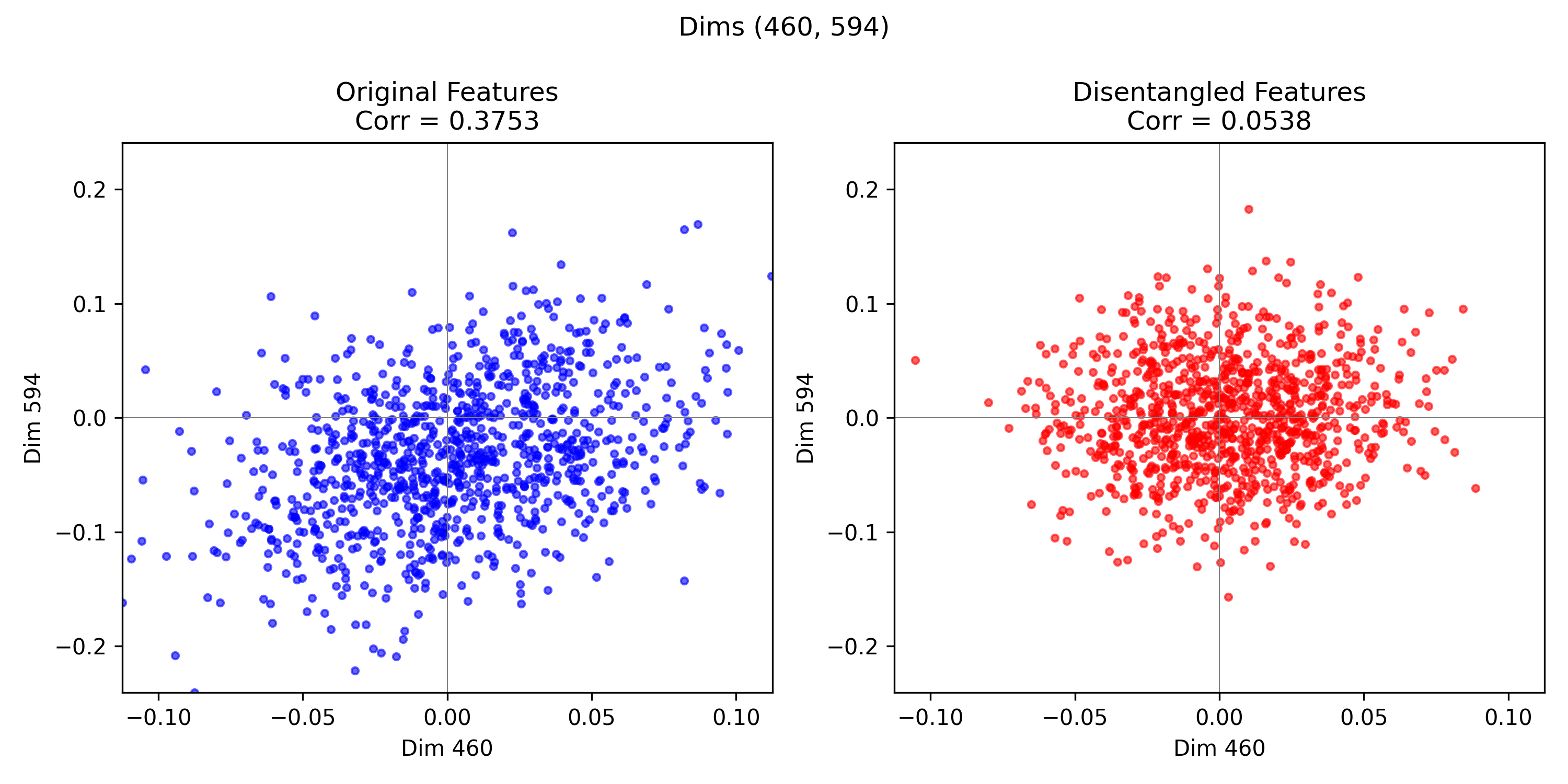}

    
    \includegraphics[width=0.6\textwidth]{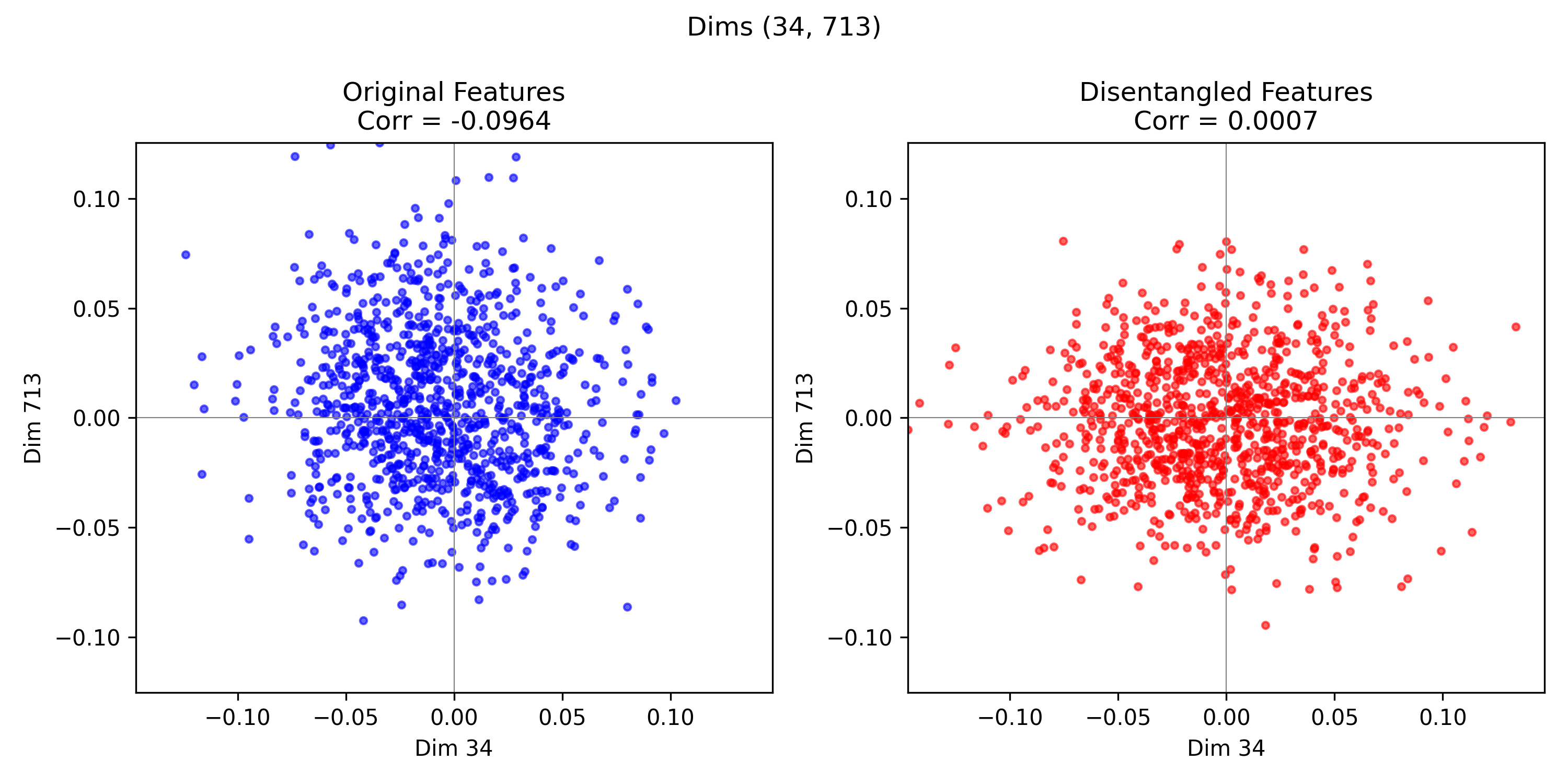}

    
    \includegraphics[width=0.6\textwidth]{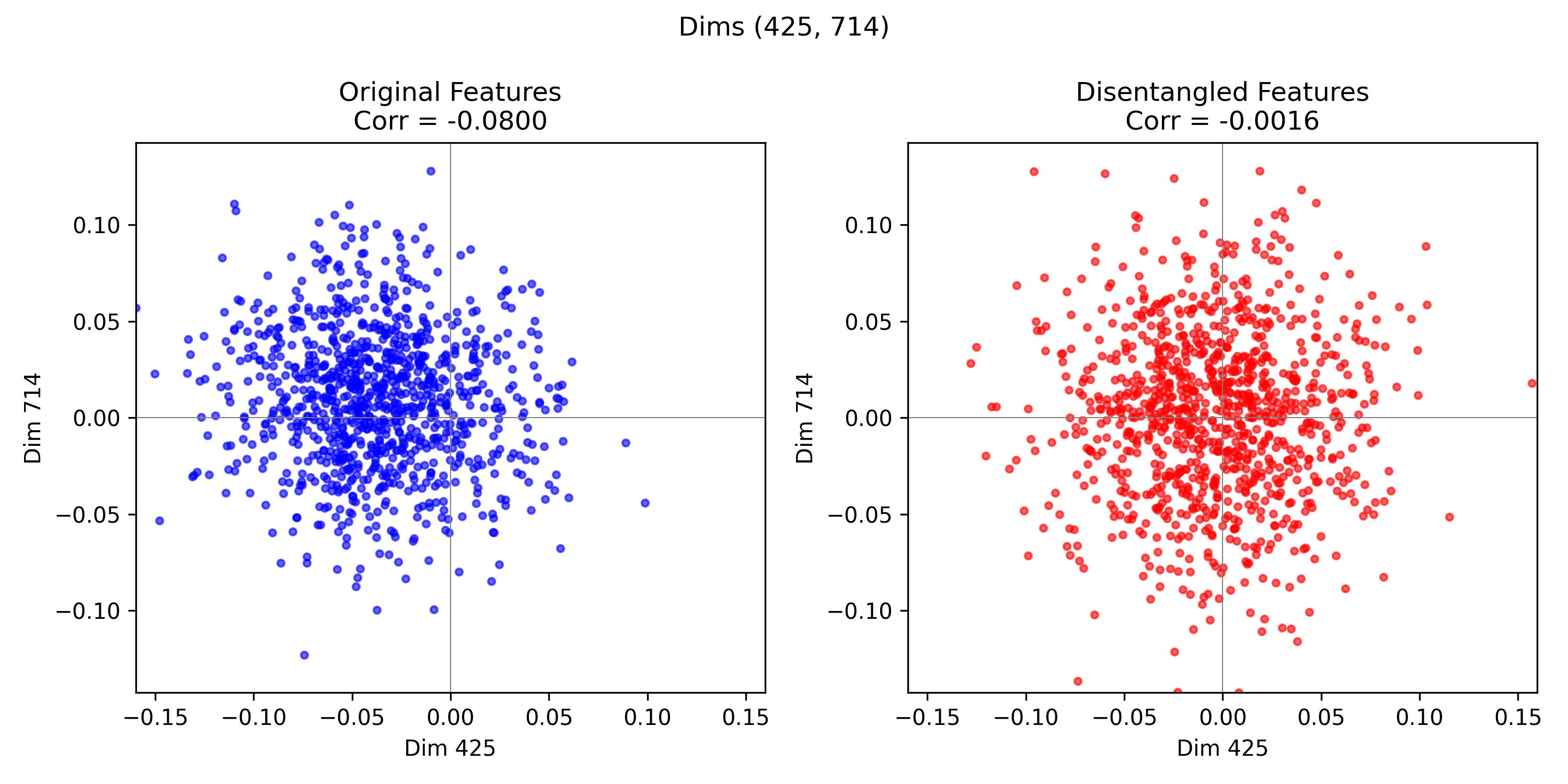}

    \caption{
\textbf{Visualization of 1000 synthetic image features before and after disentanglement using DINO as the alignment backbone.} 
Each row corresponds to a randomly selected pair of feature dimensions. 
The left column shows the original features (alignment), while the right column shows the mapped (disentangled) features. 
We report the Pearson correlation coefficient for each pair, demonstrating how the disentanglement makes the feature dimensions uncorrelated.
}
    \label{fig:random_dim_vis_vertical}
\end{figure}

\paragraph{Effect of disentanglement}
As illustrated in Figure \ref{fig:random_dim_vis_vertical}, the features clearly becomes more uncorrelated after disentanglement.

\section{Regularization Details}
\label{sec:regularization_details}

\paragraph{\( \mathbf{H}_{S}^{\top} \mathbf{H}_{E} \) for regularization}

We use $\|\mathbf{H}_{S}^{\top}\mathbf{H}_{E}-\mathbf{I}\|_{F}$ as the regularizer. As shown in Table~\ref{tab:ablation study}, it is slightly worse than the regularization in the main paper—except when DINO is used as the alignment backbone—yet remains comparable to the supervised method, supporting our theoretical analysis. Overall, when $\mathbf{H}_{S}$ and $\mathbf{H}_{E}$ are initialized as identity matrices, either the self-orthogonality regularizers $\|\mathbf{H}_{S}^{\top}\mathbf{H}_{S}-\mathbf{I}\|_{F}$ and $\|\mathbf{H}_{E}^{\top}\mathbf{H}_{E}-\mathbf{I}\|_{F}$ or the cross-term regularizer $\|\mathbf{H}_{S}^{\top}\mathbf{H}_{E}-\mathbf{I}\|_{F}$ alone is sufficient to preserve alignment.

\paragraph{Regularization matters}
As shown in Table \ref{tab:without reg}, attribution performance degrades over time when regularization is not applied, whereas with regularization, performance remains stable. The reason why our A\&D method still performs well in the early epochs without regularization is due to the identity initialization, which is an orthogonal matrix. Additionally, the initialization for the two linear mappings is identical, further preventing permutation at the beginning. However, as training progresses, the likelihood of permutation increases and the alignment breaks without regularization, leading to degradation in attribution performance.

\paragraph{Appropriate values for the hyperparameter \(\lambda\)}
We trained our A\&D method using the DINO alignment backbone with different values of \(\lambda\), which controls \(\mathcal{L}_{\text{Reg}}\). Figure \ref{fig:different_lambda} shows the resulting attribution performance. For better comparison, we also include the performance of the supervised method as dashed lines. The graph indicates that both too low and too high values of \(\lambda\) are detrimental to A\&D's performance. A lower \(\lambda\) leads to degradation over time, as discussed earlier, while a higher \(\lambda\) restricts the mappings' ability to make the features more independent. Empirically, \(\lambda\) should be chosen between 0.05 and 0.15 to balance performance and training stability.

\section{Full Experimental Results}
\label{sec:full_experimental_results}

In addition to using the object + style training set, we also conducted experiments on the object-centric training set and the style training set. The results are presented in Tables \ref{tab:Full evaluation results on ImageNet-Seen}, \ref{tab:Full evaluation results on BAM-FG.}, \ref{tab:Full evaluation results on ImageNet-Unseen.}, \ref{tab:Full evaluation results on Artchive.}. We also report Recall@10 and Recall@100 in these tables.

\begin{table*}[]
\setlength{\tabcolsep}{0.5mm}
\centering
\caption{\textbf{Performance of combining our A\&D with Supervised training and Pseudo labeling.} The models are trained on the object + style training set.}
\resizebox{\textwidth}{!}{%
\begin{tabular}{llcccccccccccccccc|cc}
\toprule
\multicolumn{2}{l}{Source} & \multicolumn{4}{c}{ImageNet-Seen} & \multicolumn{4}{c}{BAM-FG} & \multicolumn{4}{c}{ImageNet-Unseen} & \multicolumn{4}{c|}{Artchive} & 

\\ \cmidrule(lr){1-2} \cmidrule(lr){3-6} \cmidrule(lr){7-10} \cmidrule(lr){11-14} \cmidrule(lr){15-18}
\multicolumn{2}{l}{Prompts} & \multicolumn{2}{c}{GPT} & \multicolumn{2}{c}{Media} & \multicolumn{2}{c}{GPT} & \multicolumn{2}{c}{Object} & \multicolumn{2}{c}{GPT} & \multicolumn{2}{c}{Media} & \multicolumn{2}{c}{GPT} & \multicolumn{2}{c|}{Object} & \multicolumn{2}{c}{Average}\\ \cmidrule(lr){1-2} \cmidrule(lr){3-4} \cmidrule(lr){5-6} \cmidrule(lr){7-8} \cmidrule(lr){9-10} \cmidrule(lr){11-12} \cmidrule(lr){13-14} \cmidrule(lr){15-16} \cmidrule(lr){17-18}
$F_\text{base}$ & Method & R@5 & mAP & R@5 & mAP & R@5 & mAP & R@5 & mAP & R@5 & mAP & R@5 & mAP & R@5 & mAP & R@5 & mAP & R@5 & mAP \\ \midrule

\multirow{6}{*}{DINO}  
&Pretrain & 0.433 & 0.393 & 0.288 & 0.255    & 0.148 & 0.163 & 0.193 & 0.219   & 0.831 & 0.795 & 0.540 & 0.492   & 0.183 & 0.181 & 0.140 & 0.136  & 0.345 & 0.329 \\

&A\&D  & 0.491 & 0.448 & 0.364 & 0.323 & 0.167 & 0.186 & 0.210 & 0.239 & 0.845 & 0.811 & 0.633 & 0.581 & 0.212 & 0.214 & 0.164 & 0.161 & 0.386 & 0.370 \\
\cdashline{2-20}

&Pseudo & \textbf{0.471} & \textbf{0.426} & \textbf{0.374} & \textbf{0.330} & 0.127 & 0.142 & 0.164 & 0.187 & 0.840 & 0.804 & \textbf{0.639} & \textbf{0.588} & 0.211 & 0.216 & 0.161 & 0.160  & 0.373 & 0.357 \\
&A\&D+Pse & 0.466 & 0.424 & 0.352 & 0.311 & \textbf{0.165} & \textbf{0.185} & \textbf{0.213} & \textbf{0.243} & \textbf{0.842} & \textbf{0.808} & 0.608 & 0.559 & \textbf{0.225} & \textbf{0.233} & \textbf{0.178 }& \textbf{0.179} & \textbf{0.381} & \textbf{0.368} \\

\cdashline{2-20}
&Superv & 0.475 & 0.433 & 0.351 & 0.311   & \textbf{0.165} & \textbf{0.184} & \textbf{0.212} & \textbf{0.239}   & 0.842 & 0.810 & 0.598 & 0.549 & \textbf{0.217} & \textbf{0.221} & \textbf{0.170} & \textbf{0.169 }& \textbf{0.379} & \textbf{0.365}  \\

&A\&D+Sup & \textbf{0.484} &\textbf{ 0.441} &\textbf{ 0.359} & \textbf{0.320} & 0.147 & 0.158 & 0.190 & 0.211 & \textbf{0.844} & \textbf{0.811} & \textbf{0.610} & \textbf{0.562} & 0.191 & 0.194 & 0.150 & 0.149 & 0.372 & 0.356 \\

\bottomrule
\end{tabular}%
}
\label{tab:AD + pseudo and supervise}
\end{table*}

\begin{table*}[]
\setlength{\tabcolsep}{0.5mm}
\centering
\caption{\textbf{A\&D with DINO alignment backbone trained without $\lambda$.} The models are trained on the object + style training set. The table shows the performance degradation as the training progresses without $\lambda$.}
\resizebox{\textwidth}{!}{%
\begin{tabular}{ll|cccccccccccccccc|cc}
\toprule
\multicolumn{2}{l}{Source} & \multicolumn{4}{c}{ImageNet-Seen} & \multicolumn{4}{c}{BAM-FG} & \multicolumn{4}{c}{ImageNet-Unseen} & \multicolumn{4}{c|}{Artchive} 
& 

\\ \cmidrule(lr){1-2} \cmidrule(lr){3-6} \cmidrule(lr){7-10} \cmidrule(lr){11-14} \cmidrule(lr){15-18}
\multicolumn{2}{l}{Prompts} & \multicolumn{2}{c}{GPT} & \multicolumn{2}{c}{Media} & \multicolumn{2}{c}{GPT} & \multicolumn{2}{c}{Object} & \multicolumn{2}{c}{GPT} & \multicolumn{2}{c}{Media} & \multicolumn{2}{c}{GPT} & \multicolumn{2}{c|}{Object} & 
\multicolumn{2}{c}{Average} 
\\ \cmidrule(lr){1-2} \cmidrule(lr){3-4} \cmidrule(lr){5-6} \cmidrule(lr){7-8} \cmidrule(lr){9-10} \cmidrule(lr){11-12} \cmidrule(lr){13-14} \cmidrule(lr){15-16} \cmidrule(lr){17-18}
$\lambda$ & Epochs & R@5 & mAP & R@5 & mAP & R@5 & mAP & R@5 & mAP & R@5 & mAP & R@5 & mAP & R@5 & mAP & R@5 & mAP & R@5 & mAP\\ \midrule

\multirow{5}{*}{$\lambda=0$ }  

& No.10 & 0.482 & 0.446 & 0.340 & 0.309 & 0.159 & 0.177 & 0.211 & 0.240 & 0.856 & 0.825 & 0.624 & 0.579 & 0.204 & 0.205 & 0.162 & 0.159 & \textbf{0.380}	& \textbf{0.368}	 \\
& No.20 & 0.461 & 0.429 & 0.323 & 0.296 & 0.156 & 0.174 & 0.210 & 0.239 & 0.853 & 0.822 & 0.611 & 0.568 & 0.196 & 0.196 & 0.155 & 0.151 & 0.371	& 0.359\\
& No.30 & 0.454 & 0.423 & 0.319 & 0.292 & 0.156 & 0.173 & 0.210 & 0.238 & 0.852 & 0.821 & 0.605 & 0.562 & 0.196 & 0.194 & 0.153 & 0.149 & 0.368	& 0.356	\\
& No.40 & 0.452 & 0.421 & 0.317 & 0.290 & 0.155 & 0.172 & 0.209 & 0.237 & 0.851 & 0.820 & 0.604 & 0.561 & 0.195 & 0.192 & 0.153 & 0.148 & 0.367	& 0.355 \\

\midrule
\multirow{5}{*}{$\lambda=0.1$ }

& No.10 & 0.491 & 0.448 & 0.364 & 0.323 & 0.167 & 0.186 & 0.210 & 0.239 & 0.845 & 0.811 & 0.633 & 0.581 & 0.212 & 0.214 & 0.164 & 0.161 & \textbf{0.386} & \textbf{0.370}\\
& No.20 & 0.491 & 0.447 & 0.358 & 0.320 & 0.166 & 0.184 & 0.210 & 0.239 & 0.846 & 0.812 & 0.634 & 0.581 & 0.211 & 0.212 & 0.164 & 0.160 & 0.385 & 0.369\\
& No.30 & 0.491 & 0.447 & 0.358 & 0.320 & 0.164 & 0.182 & 0.209 & 0.238 & 0.847 & 0.814 & 0.634 & 0.582 & 0.210 & 0.209 & 0.162 & 0.159 & 0.384 & 0.369\\
& No.40 & 0.490 & 0.448 & 0.357 & 0.320 & 0.164 & 0.182 & 0.209 & 0.238 & 0.846 & 0.814 & 0.634 & 0.582 & 0.209 & 0.208 & 0.161 & 0.157 & 0.384 & 0.369\\

\bottomrule
\end{tabular}%
}
\label{tab:without reg}
\end{table*}

\begin{table*}[]
\setlength{\tabcolsep}{0.5mm}
\centering
\caption{\textbf{Ablation results.} Here, HSHE means the $\mathcal{L}_{\text{Reg}}$ for A\&D is calculated using $\|\mathbf{H}_{S}^{\top} \mathbf{H}_{E} - \mathbf{I}\|_F$. RandS means $\mathbf{H}_{S}$ and $\mathbf{H}_{E}$ are initialized as the same random matrix. RandD means $\mathbf{H}_{S}$ and $\mathbf{H}_{E}$ are initialized as different random matrices.}
\resizebox{\textwidth}{!}{%
\begin{tabular}{llcccccccccccccccc|cc}
\toprule
\multicolumn{2}{l}{Source} & \multicolumn{4}{c}{ImageNet-Seen} & \multicolumn{4}{c}{BAM-FG} & \multicolumn{4}{c}{ImageNet-Unseen} & \multicolumn{4}{c|}{Artchive} 
& 

\\ \cmidrule(lr){1-2} \cmidrule(lr){3-6} \cmidrule(lr){7-10} \cmidrule(lr){11-14} \cmidrule(lr){15-18}
\multicolumn{2}{l}{Prompts} & \multicolumn{2}{c}{GPT} & \multicolumn{2}{c}{Media} & \multicolumn{2}{c}{GPT} & \multicolumn{2}{c}{Object} & \multicolumn{2}{c}{GPT} & \multicolumn{2}{c}{Media} & \multicolumn{2}{c}{GPT} & \multicolumn{2}{c|}{Object} & 
\multicolumn{2}{c}{Average} 
\\ \cmidrule(lr){1-2} \cmidrule(lr){3-4} \cmidrule(lr){5-6} \cmidrule(lr){7-8} \cmidrule(lr){9-10} \cmidrule(lr){11-12} \cmidrule(lr){13-14} \cmidrule(lr){15-16} \cmidrule(lr){17-18}
$F_\text{base}$ & Method & R@5 & mAP & R@5 & mAP & R@5 & mAP & R@5 & mAP & R@5 & mAP & R@5 & mAP & R@5 & mAP & R@5 & mAP & R@5 & mAP\\ \midrule

\multirow{5}{*}{MoCo}  

&Pretrain & 0.390 & 0.347 & 0.239 & 0.211    & 0.130 & 0.144 & 0.187 & 0.211   & 0.761 & 0.717 & 0.460 & 0.408   & 0.169 & 0.164 & 0.131 & 0.125 & 0.308 & 0.291 \\
&HSHE & 0.437 & 0.393 & 0.293 & 0.260 & 0.150 & 0.168 & 0.210 & 0.240 & 0.797 & 0.758 & 0.529 & 0.479 & 0.201 & 0.201 & 0.159 & 0.158 & 0.347& 0.332	\\
&RandS & 0.434 & 0.390 & 0.291 & 0.259 & 0.149 & 0.167 & 0.210 & 0.240 & 0.796 & 0.758 & 0.529 & 0.478 & 0.201 & 0.201 & 0.159 & 0.158 & 0.346 & 0.331\\
&RandD & - & - & - & - & - & - & - & - & - & - & - & - & - & - & - & - & - & - \\
&A\&D  & 0.444 & 0.399 & 0.313 & 0.277 & 0.157 & 0.175 & 0.209 & 0.238 & 0.804 & 0.765 & 0.569 & 0.512 & 0.204 & 0.204 & 0.158 & 0.154 & \textbf{0.357} & \textbf{0.341} \\

\rowcolor{gray!40}&Supervise & 0.437 & 0.394 & 0.295 & 0.262   & 0.153 & 0.172 & 0.209 & 0.238   & 0.791 & 0.753 & 0.519 & 0.467   & 0.208 & 0.209 & 0.165 & 0.165 & 0.347 & 0.333 \\

\midrule

\multirow{5}{*}{DINO}  

&Pretrain & 0.433 & 0.393 & 0.288 & 0.255    & 0.148 & 0.163 & 0.193 & 0.219   & 0.831 & 0.795 & 0.540 & 0.492   & 0.183 & 0.181 & 0.140 & 0.136  & 0.345 & 0.329 \\
&HSHE & 0.501 & 0.459 & 0.361 & 0.325 & 0.164 & 0.184 & 0.213 & 0.246 & 0.852 & 0.820 & 0.626 & 0.578 & 0.217 & 0.221 & 0.171 & 0.170 & \textbf{0.388} & \textbf{0.375}	 \\
&RandS & 0.494 & 0.454 & 0.354 & 0.318 & 0.164 & 0.182 & 0.214 & 0.245 & 0.851 & 0.818 & 0.622 & 0.573 & 0.215 & 0.218 & 0.169 & 0.168 & 0.385 & 0.372	\\
&RandD & - & - & - & - & - & - & - & - & - & - & - & - & - & - & - & - & - & - \\
&A\&D  & 0.491 & 0.448 & 0.364 & 0.323 & 0.167 & 0.186 & 0.210 & 0.239 & 0.845 & 0.811 & 0.633 & 0.581 & 0.212 & 0.214 & 0.164 & 0.161 & 0.386 & 0.370 \\

\rowcolor{gray!40}
&Supervise & 0.475 & 0.433 & 0.351 & 0.311   & 0.165 & 0.184 & 0.212 & 0.239   & 0.842 & 0.810 & 0.598 & 0.549 & 0.217 & 0.221 & 0.170 & 0.169 & 0.379 & 0.365  \\

\midrule

\multirow{5}{*}{ViT} 
& Pretrain   & 0.355 & 0.310 & 0.242 & 0.210    & 0.168 & 0.193 & 0.224 & 0.259    & 0.785 & 0.726 & 0.474 & 0.410 & 0.201  & 0.210 & 0.156 & 0.161 & 0.326 & 0.310\\
& HSHE & 0.454 & 0.408 & 0.331 & 0.290 & 0.168 & 0.191 & 0.217 & 0.252 & 0.827 & 0.782 & 0.580 & 0.520 & 0.165 & 0.165 & 0.128 & 0.127 & 0.359	& 0.342\\
&RandS & 0.440 & 0.396 & 0.316 & 0.277 & 0.169 & 0.192 & 0.221 & 0.255 & 0.828 & 0.783 & 0.573 & 0.511 & 0.166 & 0.166 & 0.130 & 0.127 & 0.355 & 0.338\\
&RandD & - & - & - & - & - & - & - & - & - & - & - & - & - & - & - & - & - & - \\
& A\&D   & 0.440 & 0.395 & 0.295 & 0.259 & 0.203 & 0.229 & 0.251 & 0.288 & 0.845 & 0.805 & 0.561 & 0.500 & 0.229 & 0.240 & 0.176 & 0.182 & \textbf{0.375} & \textbf{0.362} \\

\rowcolor{gray!40}
& Supervise  & 0.448 & 0.398 & 0.315 & 0.274    & 0.182 & 0.209 & 0.234 & 0.272    & 0.820 & 0.772 & 0.539 & 0.474 & 0.206 & 0.216 & 0.159 & 0.163 & 0.363 & 0.347\\

\midrule
\bottomrule
\end{tabular}%
}
\label{tab:ablation study}
\end{table*}

\begin{table*}[]
\setlength{\tabcolsep}{1mm}
\centering
\caption{\textbf{Full evaluation results on ImageNet-Seen.}}
\renewcommand{\arraystretch}{0.95} 
\begin{tabular}{lllcccccccc}
\toprule
\multicolumn{3}{l}{Source}                           & \multicolumn{8}{c}{ImageNet-Seen}                             \\ \cmidrule(lr){1-3} \cmidrule(lr){4-11}
\multicolumn{3}{l}{Prompts}                          & \multicolumn{4}{c}{GPT}       & \multicolumn{4}{c}{Media}  \\ \cmidrule(lr){1-3} \cmidrule(lr){4-7} \cmidrule(lr){8-11}
$F_\text{base}$ & Method & Dataset & R@5   & R@10  & R@100 & mAP   & R@5   & R@10  & R@100 & mAP   \\ \midrule
\multirow{8}{*}{MoCo}  
        & Pretrain & None & 0.390 & 0.425 & 0.537 & 0.347 & 0.239 & 0.267 & 0.372 & 0.211 \\
    \cdashline{2-11} 
    &  & Object & 0.443 & 0.479 & 0.589 & 0.400 & 0.313 & 0.345 & 0.457 & 0.278 \\
    & Supervised & Style & 0.409 & 0.442 & 0.555 & 0.365 & 0.263 & 0.292 & 0.400 & 0.232 \\ 
    &  & Object+Style & 0.437 & 0.472 & 0.581 & 0.394 & 0.295 & 0.327 & 0.435 & 0.262 \\
    \cdashline{2-11}
    &  & Object & 0.424 & 0.456 & 0.559 & 0.381 & 0.258 & 0.285 & 0.384 & 0.227 \\
    & Pseudo & Style & 0.402 & 0.439 & 0.556 & 0.355 & 0.265 & 0.294 & 0.406 & 0.230 \\ 
    &  & Object+Style & 0.418 & 0.452 & 0.556 & 0.377 & 0.251 & 0.279 & 0.375 & 0.221 \\
    \cdashline{2-11}
    &  & Object & 0.447 & 0.477 & 0.577 & 0.404 & 0.310 & 0.340 & 0.445 & 0.276 \\
    & A\&D & Style & 0.422 & 0.457 & 0.579 & 0.377 & 0.299 & 0.334 & 0.465 & 0.264 \\ 
    &  & Object+Style & 0.444 & 0.478 & 0.589 & 0.399 & 0.313 & 0.347 & 0.468 & 0.277 \\
    \midrule

\multirow{8}{*}{DINO}  
    & Pretrain & None & 0.433 & 0.467 & 0.579 & 0.393 & 0.288 & 0.321 & 0.428 & 0.255 \\
    \cdashline{2-11} 
    &  & Object & 0.479 & 0.517 & 0.628 & 0.437 & 0.368 & 0.399 & 0.511 & 0.325 \\
    & Supervised & Style & 0.443 & 0.476 & 0.590 & 0.402 & 0.315 & 0.348 & 0.461 & 0.278 \\ 
    &  & Object+Style & 0.475 & 0.510 & 0.623 & 0.433 & 0.351 & 0.383 & 0.496 & 0.311 \\
    \cdashline{2-11}
    &  & Object & 0.463 & 0.497 & 0.606 & 0.420 & 0.333 & 0.366 & 0.479 & 0.294 \\
    & Pseudo & Style & 0.428 & 0.467 & 0.591 & 0.384 & 0.323 & 0.358 & 0.481 & 0.282 \\ 
    &  & Object+Style & 0.471 & 0.509 & 0.625 & 0.426 & 0.374 & 0.408 & 0.527 & 0.330 \\
    \cdashline{2-11}
    &  & Object & 0.493 & 0.524 & 0.620 & 0.450 & 0.354 & 0.384 & 0.484 & 0.317 \\
    & A\&D & Style & 0.452 & 0.488 & 0.607 & 0.411 & 0.348 & 0.382 & 0.509 & 0.306 \\ 
    &  & Object+Style & 0.491 & 0.523 & 0.625 & 0.448 & 0.364 & 0.395 & 0.507 & 0.323 \\
    \midrule

\multirow{8}{*}{ViT}  
    & Pretrain   & None & 0.355 & 0.393 & 0.530 & 0.310 & 0.242 & 0.275 & 0.413 & 0.210 \\
    \cdashline{2-11} 
    &  & Object & 0.452 & 0.489 & 0.615 & 0.406 & 0.335 & 0.372 & 0.509 & 0.294  \\
    & Supervised & Style & 0.357 & 0.396 & 0.539 & 0.314 & 0.253 & 0.290 & 0.432 & 0.219 \\ 
    &  & Object+Style & 0.448 & 0.487 & 0.618 & 0.398 & 0.315 & 0.353 & 0.492 & 0.274  \\
    \cdashline{2-11}
    &  & Object & 0.417 & 0.454 & 0.579 & 0.373 & 0.289 & 0.324 & 0.457 & 0.252 \\
    & Pseudo & Style & 0.370 & 0.409 & 0.543 & 0.325 & 0.257 & 0.292 & 0.428 & 0.221 \\
    &  & Object+Style & 0.361 & 0.400 & 0.539 & 0.319 & 0.266 & 0.305 & 0.447 & 0.230 \\
    \cdashline{2-11}
    &  & Object & 0.440 & 0.476 & 0.596 & 0.394 & 0.314 & 0.354 & 0.487 & 0.275 \\
    & A\&D & Style & 0.387 & 0.425 & 0.566 & 0.342 & 0.269 & 0.306 & 0.461 & 0.233 \\
    &  & Object+Style & 0.440 & 0.474 & 0.588 & 0.395 & 0.295 & 0.331 & 0.457 & 0.259 \\
    \midrule
    
\multirow{8}{*}{CLIP}  
        & Pretrain & None & 0.236 & 0.277 & 0.437 & 0.195 & 0.137 & 0.160 & 0.274 & 0.118 \\
    \cdashline{2-11} 
    &  & Object & 0.350 & 0.397 & 0.561 & 0.298 & 0.293 & 0.336 & 0.503 & 0.249 \\
    & Supervised & Style & 0.277 & 0.317 & 0.478 & 0.232 & 0.176 & 0.204 & 0.346 & 0.153 \\ 
    &  & Object+Style & 0.329 & 0.376 & 0.540 & 0.280 & 0.222 & 0.258 & 0.428 & 0.190 \\
    \cdashline{2-11}
    &  & Object & 0.110 & 0.139 & 0.273 & 0.087 & 0.071 & 0.090 & 0.192 & 0.058 \\
    & Pseudo & Style & 0.100 & 0.121 & 0.230 & 0.082 & 0.036 & 0.043 & 0.080 & 0.030 \\ 
    &  & Object+Style & 0.018 & 0.023 & 0.057 & 0.015 & 0.014 & 0.017 & 0.032 & 0.011 \\
    \cdashline{2-11}
    &  & Object & 0.286 & 0.329 & 0.493 & 0.239 & 0.183 & 0.213 & 0.358 & 0.158 \\
    & A\&D & Style & 0.239 & 0.282 & 0.443 & 0.196 & 0.142 & 0.166 & 0.289 & 0.122 \\
    &  & Object+Style & 0.257 & 0.300 & 0.468 & 0.212 & 0.153 & 0.177 & 0.304 & 0.132 \\
    \midrule

\multirow{8}{*}{SSCD}  
        & Pretrain & None & 0.253 & 0.272 & 0.335 & 0.230 & 0.142 & 0.153 & 0.194 & 0.131 \\
    \cdashline{2-11} 
    &  & Object & 0.265 & 0.284 & 0.351 & 0.241 & 0.158 & 0.170 & 0.215 & 0.144 \\
    & Supervised & Style & 0.254 & 0.273 & 0.339 & 0.228 & 0.144 & 0.155 & 0.199 & 0.132 \\ 
    &  & Object+Style & 0.268 & 0.288 & 0.357 & 0.242 & 0.156 & 0.167 & 0.214 & 0.142 \\
    \cdashline{2-11}
    &  & Object & 0.236 & 0.253 & 0.314 & 0.214 & 0.133 & 0.143 & 0.182 & 0.122 \\
    & Pseudo & Style & 0.265 & 0.286 & 0.356 & 0.239 & 0.152 & 0.165 & 0.212 & 0.138 \\ 
    &  & Object+Style & 0.267 & 0.289 & 0.360 & 0.241 & 0.154 & 0.167 & 0.215 & 0.140 \\
    \cdashline{2-11}
    &  & Object & 0.262 & 0.281 & 0.345 & 0.237 & 0.147 & 0.159 & 0.201 & 0.135 \\
    & A\&D & Style & 0.259 & 0.279 & 0.343 & 0.233 & 0.146 & 0.158 & 0.201 & 0.134 \\ 
    &  & Object+Style & 0.259 & 0.280 & 0.344 & 0.234 & 0.147 & 0.158 & 0.201 & 0.134 \\
    \bottomrule

\end{tabular}
\label{tab:Full evaluation results on ImageNet-Seen}
\end{table*}

\begin{table*}[]
\setlength{\tabcolsep}{1mm}
\centering
\caption{\textbf{Full evaluation results on BAM-FG.}}
\renewcommand{\arraystretch}{0.95} 
\begin{tabular}{lllcccccccc}
\toprule
\multicolumn{3}{l}{Source}                           & \multicolumn{8}{c}{BAM-FG}                             \\ \cmidrule(lr){1-3} \cmidrule(lr){4-11}
\multicolumn{3}{l}{Prompts}                          & \multicolumn{4}{c}{GPT}       & \multicolumn{4}{c}{Object}    \\ \cmidrule(lr){1-3} \cmidrule(lr){4-7} \cmidrule(lr){8-11}
$F_\text{base}$ & Method & Dataset & R@5   & R@10  & R@100 & mAP   & R@5   & R@10  & R@100 & mAP   \\ \midrule
\multirow{8}{*}{MoCo}  
        & Pretrain & None & 0.130 & 0.163 & 0.273 & 0.144 & 0.187 & 0.232 & 0.355 & 0.211 \\
    \cdashline{2-11} 
    &  & Object & 0.151 & 0.191 & 0.307 & 0.171 & 0.211 & 0.261 & 0.392 & 0.242 \\
    & Supervised & Style & 0.150 & 0.188 & 0.303 & 0.168 & 0.207 & 0.255 & 0.381 & 0.235 \\ 
    &  & Object+Style & 0.153 & 0.192 & 0.308 & 0.172 & 0.209 & 0.258 & 0.385 & 0.238 \\
    \cdashline{2-11}
    &  & Object & 0.130 & 0.164 & 0.269 & 0.144 & 0.185 & 0.228 & 0.347 & 0.209  \\
    & Pseudo & Style & 0.132 & 0.167 & 0.282 & 0.148 & 0.180 & 0.225 & 0.351 & 0.204 \\ 
    &  & Object+Style & 0.132 & 0.166 & 0.273 & 0.146 & 0.187 & 0.232 & 0.351 & 0.212 \\
    \cdashline{2-11}
    &  & Object & 0.146 & 0.183 & 0.292 & 0.163 & 0.206 & 0.256 & 0.382 & 0.235 \\
    & A\&D & Style & 0.157 & 0.194 & 0.301 & 0.174 & 0.206 & 0.254 & 0.371 & 0.235 \\ 
    &  & Object+Style & 0.157 & 0.195 & 0.302 & 0.175 & 0.209 & 0.257 & 0.377 & 0.238 \\
    \midrule

\multirow{8}{*}{DINO}  
    & Pretrain & None & 0.148 & 0.184 & 0.293 & 0.163 & 0.193 & 0.239 & 0.360 & 0.219 \\
    \cdashline{2-11} 
    &  & Object & 0.168 & 0.208 & 0.324 & 0.188 & 0.216 & 0.267 & 0.391 & 0.247 \\
    & Supervised & Style & 0.164 & 0.204 & 0.318 & 0.183 & 0.210 & 0.257 & 0.381 & 0.237 \\ 
    &  & Object+Style & 0.165 & 0.205 & 0.320 & 0.184 & 0.212 & 0.259 & 0.383 & 0.239 \\
    \cdashline{2-11}
    &  & Object & 0.141 & 0.176 & 0.283 & 0.156 & 0.183 & 0.226 & 0.345 & 0.207 \\
    & Pseudo & Style & 0.131 & 0.164 & 0.268 & 0.144 & 0.173 & 0.212 & 0.331 & 0.194 \\
    &  & Object+Style & 0.127 & 0.159 & 0.259 & 0.142 & 0.164 & 0.204 & 0.324 & 0.187 \\
    \cdashline{2-11}
    &  & Object & 0.161 & 0.200 & 0.312 & 0.180 & 0.212 & 0.261 & 0.387 & 0.242 \\
    & A\&D & Style & 0.164 & 0.204 & 0.309 & 0.184 & 0.205 & 0.253 & 0.369 & 0.234 \\ 
    &  & Object+Style & 0.167 & 0.206 & 0.311 & 0.186 & 0.210 & 0.259 & 0.376 & 0.239 \\
    \midrule
\multirow{8}{*}{ViT}  
    & Pretrain & None & 0.433 & 0.467 & 0.579 & 0.393 & 0.288 & 0.321 & 0.428 & 0.255 \\
    \cdashline{2-11} 
    &  & Object & 0.172 & 0.218 & 0.342 & 0.196 & 0.221 & 0.274 & 0.405 & 0.256 \\
    & Supervised & Style & 0.182 & 0.230 & 0.359 & 0.209 & 0.231 & 0.288 & 0.425 & 0.270 \\ 
    &  & Object+Style & 0.182 & 0.230 & 0.358 & 0.209 & 0.234 & 0.291 & 0.428 & 0.272  \\
    \cdashline{2-11}
    &  & Object & 0.154 & 0.195 & 0.318 & 0.174 & 0.207 & 0.259 & 0.391 & 0.239 \\
    & Pseudo & Style & 0.163 & 0.208 & 0.334 & 0.186 & 0.219 & 0.273 & 0.408 & 0.254 \\
    &  & Object+Style & 0.103 & 0.132 & 0.227 & 0.117 & 0.141 & 0.178 & 0.287 & 0.162 \\
    \cdashline{2-11}
    &  & Object & 0.167 & 0.212 & 0.336 & 0.190 & 0.219 & 0.272 & 0.404 & 0.253 \\
    & A\&D & Style & 0.187 & 0.232 & 0.343 & 0.213 & 0.227 & 0.278 & 0.392 & 0.261 \\
    &  & Object+Style & 0.203 & 0.249 & 0.355 & 0.229 & 0.251 & 0.305 & 0.418 & 0.288 \\
    \midrule
    
\multirow{8}{*}{CLIP}  
        & Pretrain & None & 0.129 & 0.170 & 0.310 & 0.148 & 0.174 & 0.225 & 0.374 & 0.200 \\
    \cdashline{2-11} 
    &  & Object & 0.166 & 0.218 & 0.364 & 0.195 & 0.216 & 0.275 & 0.431 & 0.252 \\
    & Supervised & Style & 0.185 & 0.242 & 0.405 & 0.218 & 0.235 & 0.302 & 0.472 & 0.278 \\
    &  & Object+Style & 0.186 & 0.243 & 0.408 & 0.219 & 0.236 & 0.303 & 0.474 & 0.278 \\
    \cdashline{2-11}
    &  & Object  & 0.070 & 0.097 & 0.216 & 0.084 & 0.098 & 0.129 & 0.256 & 0.115 \\
    & Pseudo & Style & 0.012 & 0.017 & 0.049 & 0.014 & 0.018 & 0.025 & 0.066 & 0.021 \\
    &  & Object+Style & 0.022 & 0.031 & 0.079 & 0.026 & 0.032 & 0.042 & 0.102 & 0.036 \\
    \cdashline{2-11}
    &  & Object & 0.150 & 0.198 & 0.346 & 0.175 & 0.196 & 0.251 & 0.407 & 0.226 \\
    & A\&D & Style & 0.162 & 0.214 & 0.365 & 0.189 & 0.205 & 0.263 & 0.417 & 0.238 \\
    &  & Object+Style  & 0.173 & 0.226 & 0.380 & 0.202 & 0.216 & 0.276 & 0.431 & 0.251 \\
    \midrule

\multirow{8}{*}{SSCD}  
        & Pretrain & None & 0.117 & 0.146 & 0.231 & 0.128 & 0.175 & 0.214 & 0.313 & 0.194 \\
    \cdashline{2-11} 
    &  & Object & 0.113 & 0.141 & 0.222 & 0.123 & 0.169 & 0.207 & 0.302 & 0.187 \\
    & Supervised & Style & 0.119 & 0.148 & 0.234 & 0.130 & 0.174 & 0.215 & 0.315 & 0.193 \\ 
    &  & Object+Style & 0.118 & 0.147 & 0.231 & 0.128 & 0.174 & 0.213 & 0.314 & 0.193 \\
    \cdashline{2-11}
    &  & Object  & 0.110 & 0.137 & 0.218 & 0.120 & 0.164 & 0.202 & 0.300 & 0.183  \\
    & Pseudo & Style & 0.117 & 0.147 & 0.231 & 0.127 & 0.170 & 0.209 & 0.309 & 0.188 \\ 
    &  & Object+Style & 0.110 & 0.137 & 0.216 & 0.118 & 0.161 & 0.198 & 0.292 & 0.178 \\
    \cdashline{2-11}
    &  & Object & 0.120 & 0.149 & 0.234 & 0.130 & 0.177 & 0.216 & 0.317 & 0.197 \\
    & A\&D & Style & 0.118 & 0.148 & 0.232 & 0.129 & 0.175 & 0.215 & 0.314 & 0.195 \\ 
    &  & Object+Style & 0.119 & 0.148 & 0.233 & 0.130 & 0.176 & 0.215 & 0.315 & 0.196 \\
    \bottomrule

\end{tabular}
\label{tab:Full evaluation results on BAM-FG.}
\end{table*}

\begin{table*}[]
\setlength{\tabcolsep}{1mm}
\centering
\caption{\textbf{Full evaluation results on ImageNet-Unseen.}}
\renewcommand{\arraystretch}{0.95}
\begin{tabular}{lllcccccccc}
\toprule
\multicolumn{3}{l}{Source}                           & \multicolumn{8}{c}{ImageNet-Unseen}                             \\ \cmidrule(lr){1-3} \cmidrule(lr){4-11}
\multicolumn{3}{l}{Prompts}                          & \multicolumn{4}{c}{GPT}       & \multicolumn{4}{c}{Media}    \\ \cmidrule(lr){1-3} \cmidrule(lr){4-7} \cmidrule(lr){8-11}
$F_\text{base}$ & Method & Dataset & R@5   & R@10  & R@100 & mAP   & R@5   & R@10  & R@100 & mAP   \\ \midrule
\multirow{8}{*}{MoCo}  
        & Pretrain & None & 0.761 & 0.786 & 0.852 & 0.717 & 0.460 & 0.493 & 0.586 & 0.408 \\
    \cdashline{2-11} 
    &  & Object & 0.792 & 0.813 & 0.874 & 0.753 & 0.542 & 0.573 & 0.658 & 0.490 \\
    & Supervised & Style & 0.783 & 0.806 & 0.868 & 0.742 & 0.493 & 0.526 & 0.612 & 0.443\\ 
    &  & Object+Style & 0.791 & 0.813 & 0.874 & 0.753 & 0.519 & 0.551 & 0.636 & 0.467 \\
    \cdashline{2-11}
    &  & Object & 0.781 & 0.805 & 0.864 & 0.741 & 0.461 & 0.494 & 0.573 & 0.411  \\
    & Pseudo & Style & 0.775 & 0.801 & 0.865 & 0.724 & 0.487 & 0.520 & 0.602 & 0.426 \\ 
    &  & Object+Style & 0.778 & 0.802 & 0.862 & 0.736 & 0.452 & 0.483 & 0.563 & 0.402 \\
    \cdashline{2-11}
    &  & Object & 0.807 & 0.828 & 0.881 & 0.769 & 0.552 & 0.584 & 0.673 & 0.501 \\
    & A\&D & Style & 0.793 & 0.818 & 0.879 & 0.754 & 0.561 & 0.601 & 0.710 & 0.502 \\ 
    &  & Object+Style & 0.804 & 0.827 & 0.882 & 0.765 & 0.569 & 0.606 & 0.709 & 0.512 \\
    \midrule

\multirow{8}{*}{DINO}  
    & Pretrain & None & 0.831 & 0.851 & 0.900 & 0.795 & 0.540 & 0.572 & 0.661 & 0.492 \\
    \cdashline{2-11} 
    &  & Object & 0.842 & 0.861 & 0.909 & 0.809 & 0.622 & 0.654 & 0.738 & 0.574 \\
    & Supervised & Style & 0.838 & 0.858 & 0.905 & 0.803 & 0.576 & 0.608 & 0.698 & 0.526 \\ 
    &  & Object+Style & 0.842 & 0.861 & 0.908 & 0.810 & 0.598 & 0.629 & 0.714 & 0.549 \\
    \cdashline{2-11}
    &  & Object & 0.838 & 0.857 & 0.904 & 0.804 & 0.583 & 0.617 & 0.705 & 0.533 \\
    & Pseudo & Style & 0.816 & 0.840 & 0.898 & 0.771 & 0.604 & 0.642 & 0.738 & 0.545 \\ 
    &  & Object+Style & 0.840 & 0.860 & 0.912 & 0.804 & 0.639 & 0.672 & 0.761 & 0.588 \\
    \cdashline{2-11}
    &  & Object & 0.849 & 0.866 & 0.907 & 0.816 & 0.622 & 0.652 & 0.734 & 0.573\\
    & A\&D & Style & 0.837 & 0.856 & 0.906 & 0.801 & 0.627 & 0.664 & 0.761 & 0.572 \\ 
    &  & Object+Style & 0.845 & 0.863 & 0.907 & 0.811 & 0.633 & 0.667 & 0.758 & 0.581 \\
    \midrule
\multirow{8}{*}{ViT}  
    & Pretrain & None & 0.785 & 0.821 & 0.898 & 0.726 & 0.474 & 0.528 & 0.689 & 0.410 \\
    \cdashline{2-11} 
    &  & Object & 0.818 & 0.844 & 0.908 & 0.773 & 0.567 & 0.615 & 0.749 & 0.505 \\
    & Supervised & Style & 0.785 & 0.820 & 0.898 & 0.727 & 0.479 & 0.534 & 0.700 & 0.415 \\ 
    &  & Object+Style & 0.820 & 0.847 & 0.912 & 0.772 & 0.539 & 0.590 & 0.737 & 0.474  \\
    \cdashline{2-11}
    &  & Object & 0.808 & 0.836 & 0.903 & 0.759 & 0.515 & 0.565 & 0.712 & 0.452 \\
    & Pseudo & Style & 0.773 & 0.810 & 0.892 & 0.712 & 0.470 & 0.527 & 0.691 & 0.407 \\
    &  & Object+Style & 0.794 & 0.827 & 0.902 & 0.737 & 0.525 & 0.580 & 0.739 & 0.457 \\
    \cdashline{2-11}
    &  & Object & 0.828 & 0.851 & 0.910 & 0.781 & 0.571 & 0.616 & 0.746 & 0.509 \\
    & A\&D & Style & 0.814 & 0.842 & 0.906 & 0.762 & 0.525 & 0.577 & 0.742 & 0.460 \\
    &  & Object+Style & 0.845 & 0.867 & 0.916 & 0.805 & 0.561 & 0.607 & 0.742 & 0.500 \\
    \midrule

\multirow{8}{*}{CLIP}  
        & Pretrain & None & 0.580 & 0.644 & 0.818 & 0.500 & 0.239 & 0.285 & 0.472 & 0.201 \\
    \cdashline{2-11} 
    &  & Object & 0.710 & 0.754 & 0.868 & 0.645 & 0.511 & 0.567 & 0.733 & 0.447 \\
    & Supervised & Style & 0.664 & 0.716 & 0.853 & 0.587 & 0.327 & 0.378 & 0.567 & 0.279 \\ 
    &  & Object+Style & 0.701 & 0.745 & 0.864 & 0.633 & 0.389 & 0.445 & 0.628 & 0.332 \\
    \cdashline{2-11}
    &  & Object  & 0.313 & 0.371 & 0.580 & 0.248 & 0.142 & 0.174 & 0.331 & 0.115 \\
    & Pseudo & Style & 0.317 & 0.371 & 0.585 & 0.259 & 0.079 & 0.093 & 0.163 & 0.064 \\
    &  & Object+Style & 0.032 & 0.042 & 0.115 & 0.027 & 0.009 & 0.011 & 0.032 & 0.007 \\
    \cdashline{2-11}
    &  & Object & 0.655 & 0.709 & 0.849 & 0.579 & 0.337 & 0.393 & 0.590 & 0.284 \\
    & A\&D & Style & 0.583 & 0.648 & 0.820 & 0.499 & 0.244 & 0.291 & 0.477 & 0.206 \\
    &  & Object+Style & 0.607 & 0.672 & 0.833 & 0.525 & 0.258 & 0.305 & 0.486 & 0.218 \\
    \midrule

\multirow{8}{*}{SSCD}  
        & Pretrain & None & 0.601 & 0.627 & 0.698 & 0.566 & 0.264 & 0.281 & 0.342 & 0.245 \\
    \cdashline{2-11} 
    &  & Object & 0.622 & 0.644 & 0.713 & 0.588 & 0.285 & 0.302 & 0.366 & 0.264 \\
    & Supervised & Style & 0.599 & 0.623 & 0.699 & 0.565 & 0.267 & 0.285 & 0.347 & 0.247 \\ 
    &  & Object+Style & 0.621 & 0.644 & 0.714 & 0.586 & 0.282 & 0.298 & 0.362 & 0.261 \\
    \cdashline{2-11}
    &  & Object  & 0.589 & 0.613 & 0.688 & 0.552 & 0.258 & 0.273 & 0.334 & 0.238  \\
    & Pseudo & Style & 0.619 & 0.643 & 0.717 & 0.581 & 0.283 & 0.301 & 0.371 & 0.260 \\
    &  & Object+Style & 0.618 & 0.642 & 0.717 & 0.581 & 0.285 & 0.304 & 0.374 & 0.263 \\
    \cdashline{2-11}
    &  & Object & 0.614 & 0.637 & 0.708 & 0.576 & 0.273 & 0.289 & 0.353 & 0.252 \\
    & A\&D & Style & 0.609 & 0.633 & 0.706 & 0.572 & 0.272 & 0.288 & 0.352 & 0.250 \\
    &  & Object+Style & 0.610 & 0.634 & 0.707 & 0.573 & 0.272 & 0.289 & 0.353 & 0.251 \\
    \bottomrule

\end{tabular}
\label{tab:Full evaluation results on ImageNet-Unseen.}
\end{table*}

\begin{table*}[]
\setlength{\tabcolsep}{1mm}
\centering
\caption{\textbf{Full evaluation results on Artchive.}}
\renewcommand{\arraystretch}{0.95} 
\begin{tabular}{lllcccccccc}
\toprule
\multicolumn{3}{l}{Source}                           & \multicolumn{8}{c}{Artchive}                             \\ \cmidrule(lr){1-3} \cmidrule(lr){4-11}
\multicolumn{3}{l}{Prompts}                          & \multicolumn{4}{c}{GPT}       & \multicolumn{4}{c}{Object}    \\ \cmidrule(lr){1-3} \cmidrule(lr){4-7} \cmidrule(lr){8-11}
$F_\text{base}$ & Method & Dataset & R@5   & R@10  & R@100 & mAP   & R@5   & R@10  & R@100 & mAP   \\ \midrule
\multirow{8}{*}{MoCo}  
        & Pretrain & None & 0.169 & 0.198 & 0.314 & 0.164 & 0.131 & 0.152 & 0.246 & 0.125 \\
    \cdashline{2-11} 
    &  & Object & 0.215 & 0.250 & 0.387 & 0.218 & 0.172 & 0.201 & 0.312 & 0.172\\
    & Supervised & Style & 0.204 & 0.237 & 0.371 & 0.204 & 0.160 & 0.188 & 0.297 & 0.160\\ 
    &  & Object+Style & 0.208 & 0.242 & 0.377 & 0.209 & 0.165 & 0.193 & 0.304 & 0.165 \\
    \cdashline{2-11}
    &  & Object & 0.162 & 0.189 & 0.297 & 0.158 & 0.130 & 0.149 & 0.233 & 0.125 \\
    & Pseudo & Style & 0.201 & 0.239 & 0.377 & 0.204 & 0.163 & 0.190 & 0.304 & 0.162 \\
    &  & Object+Style & 0.166 & 0.192 & 0.304 & 0.162 & 0.132 & 0.153 & 0.241 & 0.128 \\
    \cdashline{2-11}
    &  & Object & 0.195 & 0.224 & 0.332 & 0.193 & 0.153 & 0.173 & 0.264 & 0.150 \\
    & A\&D & Style & 0.205 & 0.234 & 0.339 & 0.204 & 0.155 & 0.177 & 0.262 & 0.152 \\
    &  & Object+Style & 0.204 & 0.234 & 0.339 & 0.204 & 0.158 & 0.180 & 0.264 & 0.154 \\
    \midrule

\multirow{8}{*}{DINO}  
    & Pretrain & None & 0.183 & 0.211 & 0.320 & 0.181 & 0.140 & 0.162 & 0.250 & 0.136 \\
    \cdashline{2-11} 
    &  & Object & 0.225 & 0.260 & 0.386 & 0.232 & 0.175 & 0.203 & 0.303 & 0.176 \\
    & Supervised & Style & 0.214 & 0.247 & 0.374 & 0.221 & 0.168 & 0.193 & 0.293 & 0.168 \\ 
    &  & Object+Style & 0.217 & 0.248 & 0.374 & 0.221 & 0.170 & 0.194 & 0.294 & 0.169 \\
    \cdashline{2-11}
    &  & Object  & 0.178 & 0.202 & 0.302 & 0.178 & 0.135 & 0.154 & 0.233 & 0.132 \\
    & Pseudo & Style & 0.157 & 0.180 & 0.271 & 0.154 & 0.119 & 0.135 & 0.207 & 0.114 \\
    &  & Object+Style  & 0.211 & 0.245 & 0.375 & 0.216 & 0.161 & 0.187 & 0.297 & 0.160 \\
    \cdashline{2-11}
    &  & Object & 0.218 & 0.249 & 0.365 & 0.222 & 0.171 & 0.196 & 0.293 & 0.171 \\
    & A\&D & Style & 0.213 & 0.239 & 0.344 & 0.214 & 0.158 & 0.178 & 0.260 & 0.156 \\
    &  & Object+Style & 0.212 & 0.240 & 0.344 & 0.214 & 0.164 & 0.184 & 0.267 & 0.161 \\
    \midrule
    
\multirow{8}{*}{ViT}  
    & Pretrain & None & 0.201 & 0.238 & 0.395 & 0.210 & 0.156 & 0.184 & 0.309 & 0.161 \\
    \cdashline{2-11} 
    &  & Object & 0.192 & 0.226 & 0.368 & 0.200 & 0.146 & 0.171 & 0.278 & 0.148 \\
    & Supervised & Style & 0.208 & 0.245 & 0.400 & 0.218 & 0.159 & 0.187 & 0.310 & 0.164 \\ 
    &  & Object+Style & 0.206 & 0.244 & 0.398 & 0.216 & 0.159 & 0.187 & 0.308 & 0.163  \\
    \cdashline{2-11}
    &  & Object & 0.157 & 0.184 & 0.305 & 0.156 & 0.128 & 0.151 & 0.250 & 0.126 \\
    & Pseudo & Style & 0.185 & 0.217 & 0.361 & 0.190 & 0.151 & 0.176 & 0.294 & 0.153 \\
    &  & Object+Style & 0.156 & 0.191 & 0.349 & 0.161 & 0.115 & 0.139 & 0.257 & 0.114 \\
    \cdashline{2-11}
    &  & Object & 0.165 & 0.193 & 0.305 & 0.165 & 0.129 & 0.150 & 0.237 & 0.127 \\
    & A\&D & Style & 0.214 & 0.247 & 0.380 & 0.226 & 0.161 & 0.185 & 0.280 & 0.166 \\
    &  & Object+Style & 0.229 & 0.264 & 0.390 & 0.240 & 0.176 & 0.202 & 0.299 & 0.182 \\
    \midrule

\multirow{8}{*}{CLIP}  
        & Pretrain & None & 0.186 & 0.234 & 0.440 & 0.198 & 0.140 & 0.175 & 0.335 & 0.144 \\
    \cdashline{2-11} 
    &  & Object & 0.249 & 0.305 & 0.512 & 0.282 & 0.188 & 0.231 & 0.396 & 0.207 \\
    & Supervised & Style & 0.264 & 0.325 & 0.565 & 0.307 & 0.203 & 0.251 & 0.447 & 0.228 \\
    &  & Object+Style & 0.259 & 0.317 & 0.550 & 0.297 & 0.201 & 0.247 & 0.437 & 0.223 \\
    \cdashline{2-11}
    &  & Object  & 0.155 & 0.203 & 0.416 & 0.178 & 0.115 & 0.150 & 0.314 & 0.126 \\
    & Pseudo & Style & 0.162 & 0.209 & 0.428 & 0.185 & 0.123 & 0.157 & 0.330 & 0.134 \\
    &  & Object+Style & 0.185 & 0.238 & 0.489 & 0.212 & 0.147 & 0.189 & 0.391 & 0.161 \\
    \cdashline{2-11}
    &  & Object & 0.214 & 0.267 & 0.481 & 0.235 & 0.163 & 0.203 & 0.372 & 0.173 \\
    & A\&D & Style & 0.207 & 0.256 & 0.463 & 0.225 & 0.156 & 0.194 & 0.352 & 0.165 \\
    &  & Object+Style & 0.229 & 0.282 & 0.500 & 0.256 & 0.176 & 0.218 & 0.387 & 0.190 \\
    \bottomrule
    
\multirow{8}{*}{SSCD}  
        & Pretrain & None & 0.151 & 0.171 & 0.241 & 0.142 & 0.123 & 0.140 & 0.203 & 0.115 \\
    \cdashline{2-11} 
    &  & Object & 0.159 & 0.179 & 0.255 & 0.151 & 0.133 & 0.149 & 0.215 & 0.125 \\
    & Supervised & Style & 0.154 & 0.173 & 0.246 & 0.144 & 0.126 & 0.142 & 0.207 & 0.118 \\ 
    &  & Object+Style & 0.159 & 0.179 & 0.254 & 0.150 & 0.131 & 0.148 & 0.213 & 0.123 \\
    \cdashline{2-11}
    &  & Object  & 0.141 & 0.159 & 0.226 & 0.132 & 0.115 & 0.131 & 0.191 & 0.107 \\
    & Pseudo & Style & 0.160 & 0.183 & 0.264 & 0.151 & 0.130 & 0.148 & 0.218 & 0.121 \\
    &  & Object+Style & 0.156 & 0.177 & 0.255 & 0.147 & 0.126 & 0.143 & 0.210 & 0.119 \\
    \cdashline{2-11}
    &  & Object & 0.154 & 0.173 & 0.246 & 0.145 & 0.125 & 0.141 & 0.205 & 0.117 \\
    & A\&D & Style & 0.150 & 0.169 & 0.239 & 0.141 & 0.122 & 0.138 & 0.201 & 0.114 \\
    &  & Object+Style & 0.150 & 0.170 & 0.241 & 0.142 & 0.123 & 0.139 & 0.202 & 0.115 \\
    \bottomrule

\end{tabular}
\label{tab:Full evaluation results on Artchive.}
\end{table*}

\end{document}